\documentclass{article} 
\usepackage{iclr2027_conference,times}

\usepackage{amsmath,amsfonts,bm}

\def\eqref#1{equation~\ref{#1}}

\def\1{\bm{1}}

\DeclareMathAlphabet{\mathsfit}{\encodingdefault}{\sfdefault}{m}{sl}
\SetMathAlphabet{\mathsfit}{bold}{\encodingdefault}{\sfdefault}{bx}{n}

\usepackage{graphicx}
\usepackage{booktabs}
\usepackage{multirow}
\usepackage{xcolor}
\usepackage{adjustbox}
\usepackage{subcaption}
\usepackage{hyperref}
\usepackage{url}
\usepackage{enumitem}
\usepackage{titletoc}
\usepackage{fontawesome5}
\usepackage{wrapfig}
\providecommand{\methodname}{\texttt{ACT\textsuperscript{3}}}

\definecolor{darkblue}{rgb}{0.0, 0.17, 0.58}
\hypersetup{colorlinks,breaklinks,linkcolor=red,citecolor=darkblue}

\usepackage{tikz,array,amssymb,graphicx,fix-cm}
\usetikzlibrary{arrows.meta,calc}

\definecolor{actSem}{HTML}{347ABB}
\definecolor{actSemFill}{HTML}{E9F2FB}
\definecolor{actVid}{HTML}{258D91}
\definecolor{actVidFill}{HTML}{E5F4F2}
\definecolor{actAct}{HTML}{D98737}
\definecolor{actActFill}{HTML}{FFF0DF}
\definecolor{actInk}{HTML}{18324D}
\definecolor{actLight}{HTML}{F2F5F7}
\definecolor{actRule}{HTML}{C8D2DB}

\newcommand{\Vtag}{%
  \begingroup\setlength{\fboxsep}{2pt}%
  \colorbox{actSemFill}{\textcolor{actSem}{\ensuremath{S}}}%
  \endgroup}

\newcommand{\Atag}{%
  \begingroup\setlength{\fboxsep}{2pt}%
  \colorbox{actActFill}{\textcolor{actAct}{\ensuremath{A}}}%
  \endgroup}

\newcommand{\Wtag}{%
  \begingroup\setlength{\fboxsep}{2pt}%
  \colorbox{actVidFill}{\textcolor{actVid}{\ensuremath{D}}}%
  \endgroup}

\definecolor{c1}{HTML}{005a32}
\definecolor{c2}{HTML}{99000d}
\newcommand{\cmark}{\textcolor{c1}{\faCheck}}
\newcommand{\xmark}{\textcolor{c2}{\faTimes}}

\title{Rewiring Semantics, Dynamics, and Control: A Simple yet Effective Action-Centric Tri-Stream Transformer}

\author{Shuang Luo$^{1}$, Yilun Kong$^{1}$, Yunpeng Qing$^{2}$, Yihang Jiao$^{2}$, Zhi Hou$^{2}$, Shunyu Liu$^{1}$, \\
\textbf{Xiaogang Wang$^{2}$, Dacheng Tao$^{1}$}\vspace{3pt} \\
$^{1}$  Nanyang Technological University, $^{2}$ ACE Robotics
}

\iclrfinalcopy 
\begin{document}

\maketitle

\begin{abstract}
Vision-Language-Action~(VLA) models have emerged as a prominent framework for complex robotic manipulation, building on the strong semantic understanding of pretrained Vision–Language Models~(VLMs). 
However, such VLM backbones offer insufficient physical dynamics priors, which limits the generalization capabilities of robot policies. 
Recent efforts therefore integrate video-generation World Models~(WMs) into robot policies through various strategies, using predictive dynamics to facilitate action generation.
Despite these advances, harnessing semantic understanding and dynamics prediction as complementary guidance for action generation remains challenging.
In this paper, we introduce \methodname, a simple yet effective \textit{\underline{A}ction-\underline{C}entric \underline{Tri}-Stream \underline{T}ransformer} that fuses semantic and dynamics information into control actions while preserving the distinct roles of context streams.
Specifically, \methodname{} enables the dedicated action expert to access VLM and WM representations through
layerwise attention, with each backbone attending only within its own stream. This straightforward interaction design maintains independent forward propagation in the context streams while allowing both backbones to be updated through control supervision.
Experiments on both simulated and real-world robotic manipulation benchmarks show that the proposed \methodname{} yields results superior to its counterparts. 
\end{abstract}

\section{Introduction}
\label{sec:introduction}

Vision-Language-Action~(VLA) models have become an established paradigm for end-to-end robotic manipulation, as exemplified by the powerful $\pi$-series models~\citep{intelligence2025pi}, which generate executable actions directly from visual observations and language instructions~\citep{brohan2023rt,kim2024openvla,black2024pi_0}.
Their capabilities build on pretrained Vision-Language Models~(VLMs), drawing on the broad semantic understanding acquired from large-scale multimodal data~\citep{beyer2024paligemma,bai2025qwen3}.
The VLM-centric architecture uses an action expert to generate actions conditioned on semantic representations from the VLM backbone~\citep{black2024pi_0,intelligence2025pi,ma2026survey}.
However, vision-language pretraining primarily emphasizes semantic understanding, providing limited physical dynamics priors for anticipating how the environment will evolve during interaction~\citep{chow2025physbench,wang2026world}. This limitation can hinder policy generalization in robotic manipulation, as effective robotic manipulation requires both understanding task semantics and accounting for the physical consequences of actions~\citep{hu2024video,wu2024unleashing,shen2026world}.

To equip robot policies with richer physical dynamics priors, recent works have explored pretrained video-generation World Models~(WMs) as policy backbones, giving rise to World Action Models~(WAMs)~\citep{cheang2024gr,kim2026cosmospolicy,ye2026dreamzero}.
The WM-centric architecture uses a video-generation backbone to support both future-frame prediction and action generation, allowing control to benefit from dynamics priors learned from large-scale video data~\citep{shen2026world,ma2026survey}.
However, directly replacing the VLM backbone with a WM may strengthen dynamics prediction at the expense of semantic understanding.
While this substitution can improve the prediction of \textit{local} state transitions, it does not necessarily preserve the task-level semantic reasoning needed to coordinate actions toward \textit{global} task objectives in robotic manipulation.

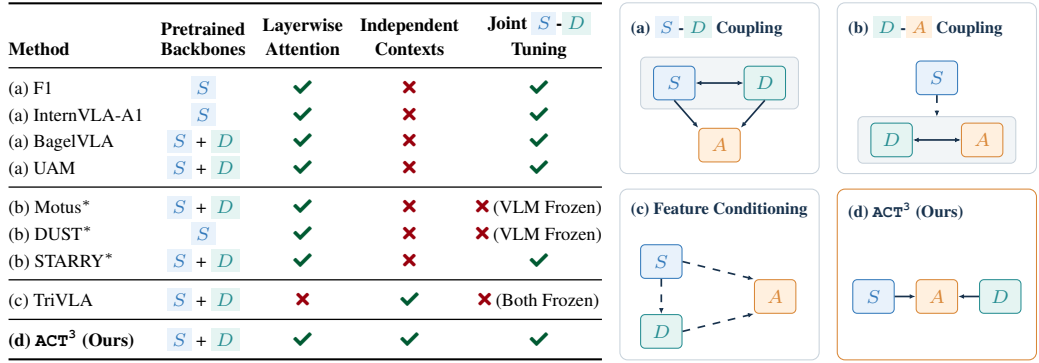
\begin{figure*}[t]
\centering
\resizebox{0.98\textwidth}{!}{%
\begin{minipage}{5.5in}
\begingroup
\setlength{\tabcolsep}{1.3pt}
\renewcommand{\arraystretch}{1.20}
\noindent
\begin{minipage}[t]{0.577\linewidth}
\vspace{0pt}
\centering
{\fontsize{7.1}{8.3}\selectfont
\def\actRowHeight{14pt}
\def\actRowDepth{5pt}
\def\actDataRowCount{9}
\def\actRowStrut{\rule[-\actRowDepth]{0pt}{\actRowHeight}}
\def\actComparisonTable{%
\begin{tabular*}{\linewidth}{@{\extracolsep{\fill}}lcccc@{}}
\toprule
\textbf{Method}
& \textbf{\shortstack{Pretrained\\Backbones}}
& \textbf{\shortstack{Layerwise\\Attention}}
& \textbf{\shortstack{Independent\\Contexts}}
& \textbf{\shortstack{Joint \Vtag-\Wtag{}\\Tuning}} \\
\midrule
\actRowStrut (a)~F1 & \Vtag{} & \cmark & \xmark & \cmark \\
\actRowStrut (a)~InternVLA-A1 & \Vtag{} & \cmark & \xmark & \cmark \\
\actRowStrut (a)~BagelVLA & \Vtag{} + \Wtag{} & \cmark & \xmark & \cmark \\
\actRowStrut (a)~UAM & \Vtag{} + \Wtag{} & \cmark & \xmark & \cmark \\
\midrule
\actRowStrut (b)~Motus$^*$ & \Vtag{} + \Wtag{} & \cmark & \xmark & \xmark{} (VLM Frozen) \\
\actRowStrut (b)~DUST$^*$ & \Vtag{} & \cmark & \xmark & \xmark{} (VLM Frozen) \\
\actRowStrut (b)~STARRY$^*$ & \Vtag{} + \Wtag{} & \cmark & \xmark & \cmark \\
\midrule
\actRowStrut (c)~TriVLA & \Vtag{} + \Wtag{} & \xmark & \cmark & \xmark{} (Both Frozen) \\
\midrule
\actRowStrut \textbf{(d)~\methodname{} (Ours)} & \Vtag{} + \Wtag{} & \cmark & \cmark & \cmark \\
\bottomrule
\end{tabular*}%
}
\setbox0=\hbox{\actComparisonTable}%
\dimen0=\dimexpr138.4pt-\ht0-\dp0\relax
\divide\dimen0 by \actDataRowCount
\edef\actRowHeight{\the\dimexpr14pt+\dimen0\relax}
\edef\actRowDepth{\the\dimexpr5pt+\dimen0/2\relax}
\actComparisonTable
}
\end{minipage}
\hfill
\begin{minipage}[t]{0.406\linewidth}
\vspace{0.2pt}
\centering
\begin{tikzpicture}[
    x=1pt,y=1pt,>={Latex[length=2.3pt,width=1.9pt]},
    box/.style={rounded corners=1.7pt,minimum width=16pt,
      minimum height=13pt,font=\fontsize{7.2}{8}\selectfont,inner sep=1pt},
    sem/.style={box,draw=actSem,fill=actSemFill,text=actSem},
    vid/.style={box,draw=actVid,fill=actVidFill,text=actVid},
    act/.style={box,draw=actAct,fill=actActFill,text=actAct},
    edge/.style={->,line width=.65pt,draw=actInk},
    both/.style={
        <->,
        line width=.65pt,
        draw=actInk
    },
    title/.style={font=\fontsize{6.5}{7.3}\selectfont\bfseries,
      anchor=north west,inner sep=0pt,text=actInk}]
\useasboundingbox (0,-138) rectangle (161,0);
\draw[actRule,rounded corners=3pt] (0,0) rectangle (77,-66);
\draw[actRule,rounded corners=3pt] (84,0) rectangle (161,-66);
\draw[actRule,rounded corners=3pt] (0,-72) rectangle (77,-138);
\filldraw[fill=white,draw=actAct,rounded corners=3pt]
    (84,-72) rectangle (161,-138);

\node[title] at (4,-5) {(a) \Vtag-\Wtag{} Coupling};
\filldraw[fill=actLight,draw=actRule,rounded corners=2pt]
    (8,-21) rectangle (69,-41);
\node[sem] (ca-v) at (21,-31) {$S$};
\node[vid] (ca-w) at (56,-31) {$D$};
\node[act] (ca-a) at (38.5,-55) {$A$};
\draw[edge] (ca-v.south) -- (ca-a.north west);
\draw[edge] (ca-w.south) -- (ca-a.north east);
\draw[both] (ca-v.east) -- (ca-w.west);

\node[title] at (88,-5) {(b) \Wtag-\Atag{} Coupling};
\node[sem] (jd-v) at (122.5,-29) {$S$};
\filldraw[fill=actLight,draw=actRule,rounded corners=2pt]
    (92,-44) rectangle (153,-62);
\node[vid] (jd-w) at (105,-53) {$D$};
\node[act] (jd-a) at (140,-53) {$A$};
\draw[edge,dashed] (jd-v.south) -- (122.5,-44);
\draw[both] (jd-w.east) -- (jd-a.west);

\node[title] at (4,-77) {(c) Feature Conditioning};
\node[sem] (fc-v) at (16,-100) {$S$};
\node[vid] (fc-w) at (16,-127) {$D$};
\node[act] (fc-a) at (60,-113.5) {$A$};
\draw[edge,dashed] (fc-v.east) -- (fc-a.north west);
\draw[edge,dashed] (fc-w.east) -- (fc-a.south west);
\draw[edge,dashed] (fc-v.south) -- (fc-w.north);

\node[title] at (88,-77) {(d) \methodname{} (Ours)};
\node[sem] (ours-v) at (98,-113.5) {$S$};
\node[act] (ours-a) at (122.5,-113.5) {$A$};
\node[vid] (ours-w) at (147,-113.5) {$D$};
\draw[edge] (ours-v) -- (ours-a);
\draw[edge] (ours-w) -- (ours-a);
\end{tikzpicture}
\end{minipage}
\endgroup
\end{minipage}%
}
\caption{Architectural comparison of different multi-stream methods.
\Vtag, \Atag, and \Wtag{} denote the semantic, action, and dynamics streams, respectively.
``$\longrightarrow$'' denotes direct cross-stream attention, whereas ``$\dashrightarrow$'' denotes conditioning on upstream features. \colorbox{gray!12}{Grey boxes} containing ``$\longleftrightarrow$'' indicate cross-stream dependencies among the grouped streams, while the exact attention direction varies across methods.
Pretrained backbones indicate which context streams are initialized from pretrained foundation models.
Layerwise attention indicates direct cross-stream interaction at multiple transformer layers, rather than using upstream features only as conditioning inputs.
Independent contexts indicates that neither context stream (\textit{i.e.}, \Vtag{} or \Wtag{}) reads from other streams during forward computation. Joint S-D tuning indicates that both context streams are updated during downstream policy training.
$^*$Motus adds a semantic expert to joint D-A attention; DUST conditions D-A denoising on frozen VLM features, while STARRY uses a trainable semantic expert.
}
\label{fig:architecture_comparison}
\end{figure*}

Recent methods further combine semantic understanding and dynamics prediction within a shared backbone~\citep{zhang2025upvla,zhao2025cotvla,cen2025worldvla,wang2026unified,chen2026unified} or specialized multi-stream architectures~\citep{lv2025f1,liu2025trivla,won2026dust,bi2026motus,cai2026internvla,hu2026bagelvla,tian2026starry,zhang2026uam}.
As summarized in Fig.~\ref{fig:architecture_comparison}, these multi-stream designs differ in how semantic and dynamics information reaches the action expert.
However, these designs often introduce cross-stream dependencies beyond the action pathway, coupling semantic and dynamics representations before or alongside action generation.
Such additional coupling is not necessarily beneficial for control, motivating a simpler alternative that integrates both sources of information directly through the action expert.

In this paper, we introduce \methodname{}, a simple yet effective \textit{\underline{A}ction-\underline{C}entric \underline{Tri}-Stream \underline{T}ransformer}.
The proposed \methodname{} adopts a Mixture-of-Transformers~(MoT) architecture, in which the VLM provides semantic understanding context for global task objectives, the WM provides predictive dynamics context for local state transitions, and the action expert serves as their sole integration point for control via action-centric cross-stream attention.
Technically, each context backbone attends only within its own stream, while action queries access both context streams through layerwise attention.
This allows each backbone to build on its pretrained modeling capabilities without incorporating representations from the other context stream, with the aim of reducing potential interference induced by cross-stream conditioning.
The action expert, in turn, learns how to weight and combine semantic and dynamics information as its action representation evolves.
We jointly optimize dynamics prediction and action generation, with action supervision also updating both context backbones through these attention connections.
This simple organization makes both the context representations and their integration learnable under the control objective, bringing global task semantics and local state dynamics together through action generation.

To sum up, our main contribution is therefore a dedicated \emph{Action-Centric Tri-Stream Transformer}. At its core, \methodname{} learns to integrate semantic understanding and predictive dynamics as complementary guidance for action generation through layerwise action queries, preserving independent context computation while enabling joint optimization for control. Notably, the proposed \methodname{} allows us to directly reuse the strong pretrained $\pi_{0.5}$ model and augment it with a pretrained Cosmos world model, providing a simple yet effective way to further improve the $\pi_{0.5}$ baseline. Experiments on various simulated and real-world robot manipulation benchmarks demonstrate the effectiveness of \methodname{}, with controlled ablations supporting the benefits of action-centric interaction. 

\section{Related Works}
\label{sec:related_work}

\textbf{Vision-Language-Action~(VLA)} models extend pretrained VLMs to robotic control, leveraging broad semantic knowledge from large-scale multimodal pretraining to generate executable actions~\citep{brohan2023rt,kim2024openvla}. Representative models from the $\pi$ series exemplify this VLM-centric paradigm~\citep{black2024pi_0,intelligence2025pi,intelligence2025pi06,intelligence2026pi07}. Recent advances have focused on open-world generalization~\citep{kim2024openvla,intelligence2025pi,li2025hamster}, improving generative action modeling~\citep{black2024pi_0,chen2026unified,hou2025dita}, and incorporating reasoning and memory mechanisms for complex manipulation~\citep{zhao2025cotvla,zhong2026acot,shi2026memoryvla}. Nevertheless, VLM-centric policies inherit limited physical dynamics priors from vision-language pretraining~\citep{chow2025physbench,hu2024video,wu2024unleashing}.

\textbf{World Action Models~(WAMs)} pursue a complementary direction by replacing the VLM-centric policy backbone with video-generation WMs, enabling control to exploit predictive dynamics learned from large-scale video data~\citep{cheang2024gr,kim2026cosmospolicy,ye2026dreamzero}. However, stronger dynamics modeling does not necessarily preserve the task-level semantic reasoning required for goal-directed manipulation, motivating policies that retain both semantic and dynamics information~\citep{yang2026world,ma2026internvla}. Existing approaches broadly adopt either a \emph{shared backbone} that jointly models understanding, prediction, and action~\citep{zhang2025upvla,zhao2025cotvla,cen2025worldvla,wang2026unified,chen2026unified}, or specialized \emph{multi-stream} architectures that maintain separate semantic, dynamics, and action pathways~\citep{lv2025f1,liu2025trivla,won2026dust,bi2026motus,cai2026internvla,hu2026bagelvla,tian2026starry,zhang2026uam}. Our work belongs to the latter family, focusing specifically on how information should flow among specialized semantic, dynamics, and action streams.

As summarized in Fig.~\ref{fig:architecture_comparison}, existing multi-stream methods mainly follow three interaction patterns: \emph{semantic-dynamics coupling} allows the semantic and dynamics representations to interact before action generation (F1~\citep{lv2025f1}, InternVLA-A1~\citep{cai2026internvla}, BagelVLA~\citep{hu2026bagelvla}, UAM~\citep{zhang2026uam}); \emph{dynamics-action coupling} tightly integrates predictive dynamics with action generation under semantic guidance (Motus~\citep{bi2026motus}, DUST~\citep{won2026dust}, STARRY~\citep{tian2026starry}); and \emph{feature conditioning} provides separate context representations as inputs to the action module (TriVLA~\citep{liu2025trivla}). In contrast, \methodname{} makes the action stream the sole integration interface: semantic and dynamics backbones remain independent during forward computation, while action queries access both through layerwise attention, allowing both backbones to be jointly optimized under action supervision.

\section{Method}
\label{sec:method}

In this section, we present \methodname{}, an action-centric tri-stream
transformer for combining semantic understanding and predictive dynamics
in robot control. The architecture consists of semantic, dynamics, and
action streams, with the action stream serving as the interface between
the two context pathways. We first introduce the problem formulation and
overall architecture (Section~\ref{sec:problem}), then describe the
layerwise action-centric interaction (Section~\ref{sec:tristream}).
Finally, we present the joint training objective and inference procedure
(Section~\ref{sec:joint_learning}).

\subsection{Overview}
\label{sec:problem}

\textbf{Problem Formulation.}
We learn an embodied policy $\pi_{\Theta}$ with trainable parameters $\Theta$ from a robot demonstration dataset $\mathcal{D}$. At control step $t$, the policy receives $M$ camera observations $\mathcal{O}_t=\{I_t^m\}_{m=1}^{M}$, where $I_t^m$ denotes the image
from camera $m$, together with a task language instruction $\ell$.
Proprioceptive state can optionally be included as an additional
conditioning input; we omit this optional input from the notation
for simplicity.
The policy models a conditional distribution over short-horizon
robot motions and predicts an $H$-step action chunk
$\hat A_t\sim\pi_{\Theta}(\cdot\mid\mathcal{O}_t,\ell)$.
We denote the corresponding ground-truth action chunk by
$A_t=[a_t,\ldots,a_{t+H-1}]\in\mathbb{R}^{H\times B}$, where
$B$ denotes the action dimension used by the policy model.
During execution, the robot executes a prefix of the predicted action chunk and replans using the updated visual observations.

\begin{figure*}[t]
    \centering
    \includegraphics[width=1.0\textwidth]{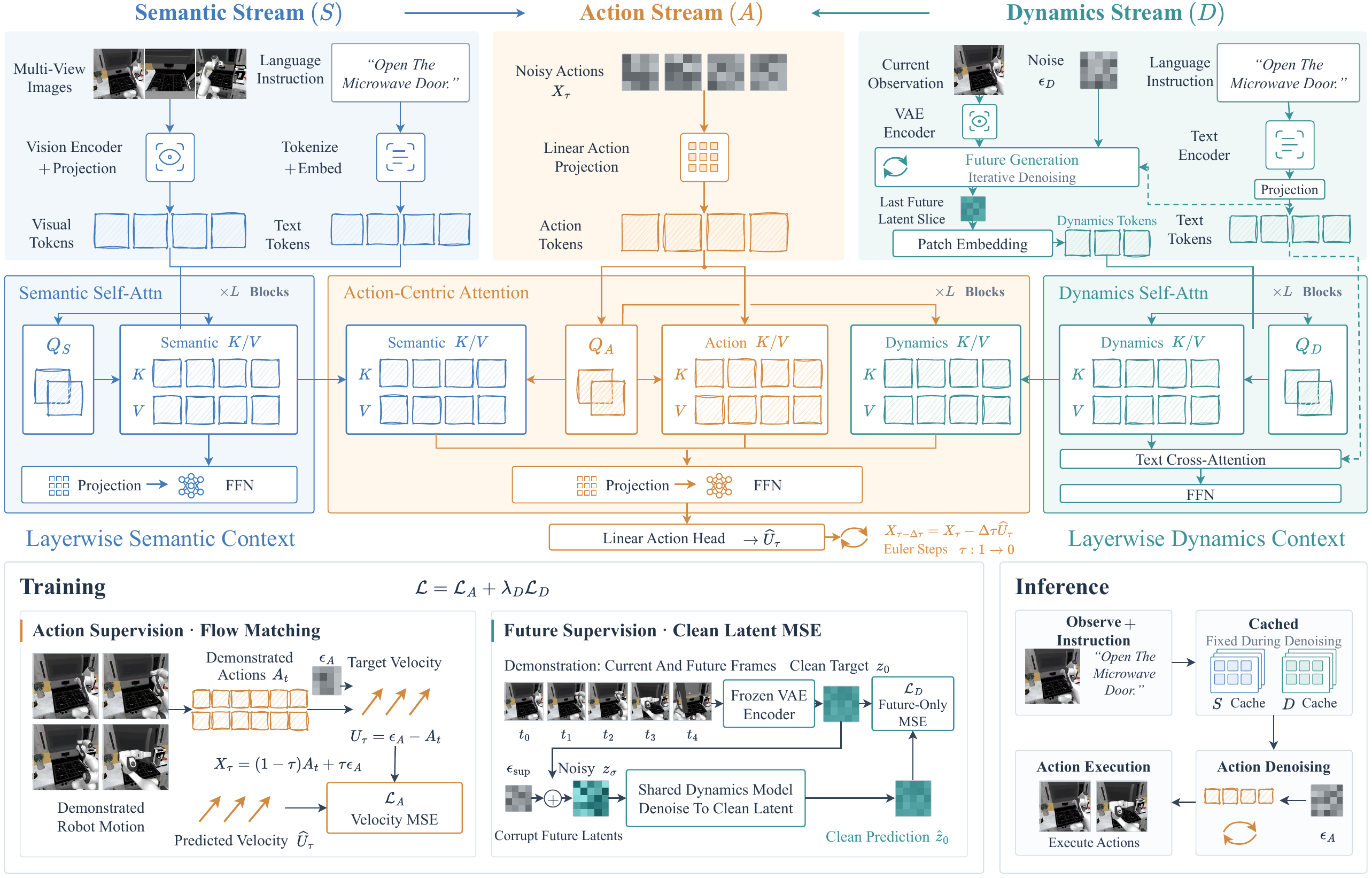}
    \caption{
    Overview of the proposed \methodname{}.
    The architecture comprises three specialized transformer streams for semantic understanding, action generation, and dynamics prediction.
    The semantic and dynamics streams maintain independent forward computation, while the action expert attends to semantic, action, and dynamics representations at every action transformer block, making the action stream the sole integration point for control.
    }
    \vspace{-0.2cm}
    \label{fig:act3_architecture}
\end{figure*}

\textbf{Model Architecture.}
As shown in Fig.~\ref{fig:act3_architecture}, \methodname{} adopts a Mixture-of-Transformers~(MoT) architecture that combines semantic understanding and predictive dynamics for control through three transformer streams. The semantic stream encodes the current multi-view observations and instruction to provide semantic context for global task objectives, while the dynamics stream provides context for anticipated local state transitions. The action stream generates continuous action chunks through flow matching. At every action transformer block, action queries attend to both semantic and dynamics tokens, while neither context stream reads representations from the other streams. This makes the action stream the sole integration point for control, combining both contexts at multiple depths while maintaining independent forward computation in the two context streams. Let $\psi$, $\theta$, and $\phi$ denote the trainable parameters of the semantic, action, and dynamics streams, respectively, with $\Theta=(\psi,\theta,\phi)$. All three streams are jointly optimized for control, as detailed in Section~\ref{sec:joint_learning}.
This organization provides a simple way to extend a pretrained MoT-based VLA by initializing the semantic stream from its VLM backbone and the action stream from its action expert, while initializing the dynamics stream from a pretrained WM.

\subsection{Action-Centric Tri-Stream Transformer}
\label{sec:tristream}

\textbf{Semantic and Dynamics Context Construction.}
Given the current observation and instruction, the semantic stream encodes the multi-view images $\mathcal{O}_t$ into visual tokens through a vision encoder, and tokenizes and embeds the instruction $\ell$ into text tokens.
These tokens are processed by the pretrained VLM transformer to provide layerwise semantic context for the action stream.
In parallel, the dynamics stream predicts future latents conditioned on the base-camera image $I_t^b$, where $b\in\{1,\ldots,M\}$, and the same instruction $\ell$.
A VAE encodes the observation, and a separate text encoder produces text tokens.
Starting from noise $\epsilon_D$ with independent standard Gaussian entries, the dynamics model samples future latents through iterative denoising conditioned on these inputs, keeping the observed latent region fixed.
We take the last temporal slice of the sampled future latents as a denoised predicted latent input and apply the dynamics patch embedding to obtain dynamics tokens.
This latent slice represents a temporally compressed future segment.
The future latents used to construct the dynamics tokens are generated using the same sampling procedure during training and inference, with $\epsilon_D$ drawn independently of the demonstrated future observations.
Demonstrated futures are used only for auxiliary prediction supervision.

\textbf{Action-Centric Attention.}
We interleave the three streams over $L$ transformer layers so that the action expert can access semantic and dynamics context as its action tokens are updated.
Only action queries read keys and values from other streams.
At layer $k=0,\ldots,L-1$, each stream computes queries $\mathbf{Q}_i^k$, keys $\mathbf{K}_i^k$, and values $\mathbf{V}_i^k$ using its own normalization and projection layers, where $i\in\{S,A,D\}$ indexes the semantic, action, and dynamics streams, respectively.
We concatenate the queries along the token dimension as $\mathbf{Q}^k=[\mathbf{Q}_S^k;\mathbf{Q}_A^k;\mathbf{Q}_D^k]$, and construct $\mathbf{K}^k$ and $\mathbf{V}^k$ analogously.
For each attention head with dimension $d$, the attention output $\mathbf{O}^k$ is
\begin{equation}
\mathbf{O}^k
=\operatorname{softmax}\!\left(
\frac{\mathbf{Q}^k(\mathbf{K}^k)^\top}{\sqrt{d}}
+\mathcal{M}
\right)\mathbf{V}^k,
\qquad
\mathcal{M}=
\begin{bmatrix}
0 & -\infty & -\infty \\
0 & 0 & 0 \\
-\infty & -\infty & 0
\end{bmatrix}.
\label{eq:joint_attention}
\end{equation}
The additive block attention mask $\mathcal{M}$ specifies stream-level visibility, with query rows and key columns ordered as $(S,A,D)$.
Each entry applies to the corresponding token block: $0$ allows attention and $-\infty$ masks it.
Action queries therefore attend jointly to semantic, action, and dynamics keys and values under a single softmax; self-attention in each context stream is restricted to that stream.
Each stream retains its own output projection and feed-forward module; the dynamics stream additionally cross-attends to its text tokens.
This asymmetric design makes the action expert the sole integration point for semantic and dynamics information and preserves independent forward computation in both context streams.

\subsection{Training and Inference}
\label{sec:joint_learning}

\textbf{Action Supervision.}
We train the policy to generate action chunks conditioned on semantic and dynamics context using continuous-action flow matching. For standard Gaussian action noise $\epsilon_A$ with the same shape as $A_t$ and flow time $\tau\sim p$ over $(0,1]$, we construct $X_\tau=(1-\tau)A_t+\tau\epsilon_A$ with target velocity $U_\tau=\epsilon_A-A_t$. Here $p$ denotes the training-time distribution over flow times. Let $\mathcal{C}_{i,t}=\{(\mathbf{K}_i^k,\mathbf{V}_i^k)\}_{k=0}^{L-1}$ collect the layerwise context for $i\in\{S,D\}$. The action expert $f_{\theta}$ linearly projects $X_\tau$ into action tokens and uses $\tau$ to condition its transformer blocks. A linear output head predicts the action velocity $\widehat U_\tau$. We minimize the action loss $\mathcal{L}_{A}$:
\begin{equation}
    \begin{aligned}
        \mathcal{L}_{A}
        =\mathbb{E}_{\mathcal{D},\tau,\epsilon_A,\epsilon_D}
        \left[
            \frac{\|\widehat U_\tau-U_\tau\|_F^2}{HB}
        \right], \qquad
        \widehat U_\tau
        =f_{\theta}(X_\tau,\tau;
                     \mathcal{C}_{S,t},\mathcal{C}_{D,t}).
    \end{aligned}
    \label{eq:action_loss}
\end{equation}
Here $\|\cdot\|_F$ denotes the Frobenius norm, and the expectation is over demonstration samples from $\mathcal{D}$ and the sampled flow times and noises.

\textbf{Joint Optimization.}
We supplement action supervision with an auxiliary dynamics-prediction loss. The same VAE encodes demonstration clips containing current and future frames into clean latents $z_0$. We corrupt their future region at a sampled noise level $\sigma$ using standard Gaussian noise $\epsilon_{\mathrm{sup}}$ to obtain noisy latents $z_\sigma$. Conditioned on the current observation and instruction, the auxiliary branch reconstructs clean latents $\hat z_0$ from $z_\sigma$. Future generation, dynamics context extraction, and auxiliary prediction share the dynamics-transformer parameters $\phi$, but the auxiliary branch provides no input to the action expert. Its loss $\mathcal{L}_{D}$ is the mean squared error between $\hat z_0$ and $z_0$ over future latent entries only. We jointly optimize all three streams under the objective $\mathcal{L}$:
\begin{equation}
    \mathcal{L}
    =\mathcal{L}_{A}
     +\lambda_{D}\mathcal{L}_{D},
    \label{eq:total_loss}
\end{equation}
where $\lambda_{D}\geq0$ weights dynamics-prediction loss. Action gradients update both context backbones through the keys and values read by the action stream. The auxiliary loss additionally updates the dynamics stream. Thus, both context backbones and the action expert are optimized for control.

\textbf{Inference with Cached Contexts.}
At each policy query, we generate future latents from the current observation and instruction and compute the layerwise contexts $\mathcal{C}_{S,t}$ and $\mathcal{C}_{D,t}$ once. These contexts remain fixed during action generation because they do not depend on the noisy actions $X_\tau$ or their flow time $\tau$. Starting from a standard Gaussian chunk $X_1$, we integrate the predicted action velocity from $\tau=1$ to $\tau=0$ using Euler steps of size $\Delta\tau>0$ to obtain $\hat A_t=X_0$. Only the action stream is recomputed during integration; both context caches are refreshed for the next policy query using updated observations. Appendix~\ref{app:method_details} provides the detailed sampling, prediction-supervision, and action-integration equations.

\section{Experiments}
\label{sec:experiments}

We organize our experiments around three questions:
(1) Does the complete policy improve task execution and robustness?
We evaluate simulated and real-world manipulation, including transfer under distribution shifts.
(2) How should actions access semantic and dynamics context?
We study cross-stream interaction and layerwise context access.
(3) What do generated context and predictive supervision contribute?
We ablate generated context and dynamics-prediction supervision.
These questions are addressed in Sections~\ref{sec:simulation_results} and~\ref{sec:real_world}, Section~\ref{sec:interaction_analysis}, and Section~\ref{sec:context_analysis}, respectively.

\subsection{Experimental Setup}

\textbf{Benchmarks and Data.}
We train one policy on the 24 kitchen tasks of RoboCasa~\citep{nasiriany2024robocasa}
and another jointly on the 40 tasks of LIBERO-Spatial, Object, Goal,
and Long~\citep{liu2023libero}. Both use a nominal budget of 50 demonstrations
per task. Following the preprocessing of OpenVLA~\citep{kim2024openvla}
for LIBERO and the successful human-demonstration subset used by
Cosmos Policy~\citep{kim2026cosmospolicy} for RoboCasa, the converted datasets
contain 1,693 and 1,199 episodes, respectively.
We further evaluate the LIBERO-trained policies on LIBERO-Plus~\citep{fei2025libero}
without additional fine-tuning, using the same 10,030 perturbed instances
for both policies. Dataset and evaluation details appear in
Appendix~\ref{app:data}, \ref{app:evaluation}, and~\ref{app:libero_plus}.

\textbf{Models and Comparisons.}
\methodname{} initializes its semantic and action streams from
$\pi_{0.5}$~\citep{intelligence2025pi} and its dynamics stream from
the Cosmos-Predict2.5-2B post-trained video checkpoint~\citep{cosmos_predict25}.
Model initialization, update scopes, and gradient-flow details are
provided in Appendix~\ref{app:model}.
The default model predicts four consecutive future images and uses generated future representations as action context
during both training and deployment. Demonstrated future observations
provide auxiliary supervision, rather than inputs to the deployed policy.

Our primary baseline is $\pi_{0.5}$ trained and evaluated in the same
PyTorch pipeline, following the official implementation~\citep{intelligence2025pi}. 
For LIBERO, the comparisons include
$\pi_0$~\citep{black2024pi_0},
OpenVLA-OFT~\citep{kim2025fine},
DUST~\citep{won2026dust},
HAMLET~\citep{koo2026hamlet},
ForeAct~\citep{zhang2026foreact},
Fast-WAM~\citep{yuan2026fast},
LingBot-VA~\citep{li2026lingbotva}, and
Cosmos Policy~\citep{kim2026cosmospolicy}.
For RoboCasa, the comparisons include
UVA~\citep{li2025uva},
DUST~\citep{won2026dust},
UWM~\citep{zhu2025uwm},
$\pi_0$~\citep{black2024pi_0},
GR00T-N1.5~\citep{bjorck2025gr00t},
FLARE~\citep{zheng2025flare},
HAMLET~\citep{koo2026hamlet}, and
Cosmos Policy~\citep{kim2026cosmospolicy}.
Checkpoint selection and the shared evaluation protocol are documented
in Appendix~\ref{app:optimization} and~\ref{app:evaluation}.

\subsection{Task Execution and Robustness}
\label{sec:simulation_results}
\label{sec:coverage}

\textbf{RoboCasa.}
\methodname{} achieves 71.2\% success on the 24 kitchen tasks, compared
with 59.4\% for the in-house $\pi_{0.5}$ baseline
(Table~\ref{tab:robocasa_main}). With the same downstream demonstrations
and available camera views, augmenting the VLA with the dynamics stream
and its training objective improves kitchen manipulation performance.
This establishes the benefit of the complete policy; the following
ablations examine how its streams and objectives contribute.

\begin{table*}[t]
    \centering

    \begin{minipage}[t]{0.58\textwidth}
        \caption{LIBERO results.
        Success rates (\%) across the Spatial, Object, Goal,
        and Long task suites.
        The last two rows use our PyTorch pipeline;
        other scores are taken from prior work under
        the original evaluation settings.}
        \label{tab:libero_main}
    \end{minipage}\hfill
    \begin{minipage}[t]{0.39\textwidth}
        \caption{RoboCasa results.
        Success rates (\%) on 24 kitchen tasks.
        The last two rows use our pipeline;
        other scores are from prior work.}
        \label{tab:robocasa_main}
    \end{minipage}
    \par\vspace{0mm}

    \begin{minipage}[t]{0.58\textwidth}
        \centering
        \small
        \setlength{\tabcolsep}{2pt}
        \renewcommand{\arraystretch}{1.1}
        \begin{tabular*}{\linewidth}{@{\extracolsep{\fill}}lccccc@{}}
            \toprule
            Method & Spatial & Object & Goal & Long & Average \\
            \midrule
            $\pi_0$
                & 96.8 & 98.8 & 95.8 & 85.2 & 94.2 \\
            OpenVLA-OFT
                & 97.6 & 98.4 & 97.9 & 94.5 & 97.1 \\
            DUST
                & 96.2 & 99.8 & 96.0 & 92.6 & 96.2 \\
            HAMLET
                & 99.0 & 100.0 & 99.2 & 92.2 & 97.6 \\
            ForeAct
                & 97.3 & 99.8 & 97.3 & 95.4 & 97.5 \\
            Fast-WAM
                & 98.2 & 100.0 & 97.0 & 95.2 & 97.6 \\
            LingBot-VA
                & 98.5 & 99.6 & 97.2 & 98.5 & 98.5 \\
            Cosmos Policy
                & 98.1 & 100.0 & 98.2 & 97.6 & 98.5 \\
            \midrule
            $\pi_{0.5}$ (PyTorch)
                & 98.0 & 98.4 & 97.5 & 93.5 & 96.9 \\
            \textbf{\boldmath\methodname}~(Ours)
                & 99.5 & 99.0 & 98.5 & 97.4 & 98.6 \\
            \bottomrule
        \end{tabular*}
    \end{minipage}\hfill
    \begin{minipage}[t]{0.39\textwidth}
        \centering
        \small
        \setlength{\tabcolsep}{1pt}
        \renewcommand{\arraystretch}{1.1}
        \begin{tabular*}{\linewidth}{@{\extracolsep{\fill}}lcc@{}}
            \toprule
            Method & \# Demos per Task & Average \\
            \midrule
            UVA
                & 50 & 50.0 \\
            DUST
                & 300 & 58.5 \\
            UWM
                & 1,000 & 60.8 \\
            $\pi_0$
                & 300 & 62.5 \\
            GR00T-N1.5
                & 300 & 64.1 \\
            FLARE
                & 300 & 66.4 \\
            HAMLET
                & 300 & 66.4 \\
            Cosmos Policy
                & 50 & 67.1 \\
            \midrule
            $\pi_{0.5}$ (PyTorch)
                & 50 & 59.4 \\
            \textbf{\boldmath\methodname}~(Ours)
                & 50 & 71.2 \\
            \bottomrule
        \end{tabular*}
    \end{minipage}
    \vspace{-0.4cm}
\end{table*}

\textbf{LIBERO.}
\methodname{} improves all four LIBERO suites, reaching a 98.6\% average
success rate from a strong 96.9\% baseline (Table~\ref{tab:libero_main}).
The improvement spans spatial, object, and goal variation as well as
long-horizon tasks. In particular, the Long result shows that repeatedly
refreshing short-range dynamics context is compatible with improved
execution over longer task sequences.

\begin{wrapfigure}{r}{0.50\linewidth}
    \centering
    \vspace{-1.2cm}
    \includegraphics[width=\linewidth]
    {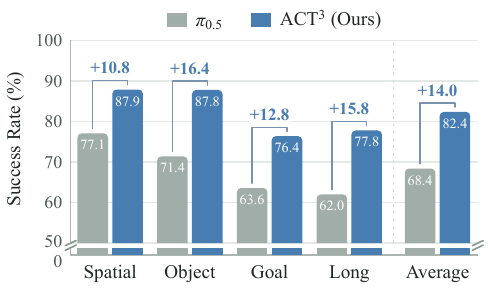}
    \caption{LIBERO-Plus success rates (\%) on 10,030 evaluation
instances under direct transfer from LIBERO without additional
fine-tuning. Both models use PyTorch implementations.}
    \label{tab:libero_plus_main}
    \vspace{-8pt}
\end{wrapfigure}

\paragraph{Transfer under Distribution Shifts.}
On LIBERO-Plus, \methodname{} achieves an average success rate of
82.4\%, compared with 68.4\% for the in-house $\pi_{0.5}$ baseline,
when both LIBERO-trained policies are evaluated on the same
10,030 instances without additional fine-tuning
(Figure~\ref{tab:libero_plus_main}).
The improvement spans all four task suites, with the largest gains
on Object and Long,
corresponding to increases of 16.4 and 15.8 percentage points,
respectively.
By integrating task semantics with predicted local state changes
at each action layer, \methodname{} may better handle object
variation and limit error accumulation over longer task sequences.
While both policies already perform strongly on standard LIBERO,
the larger performance gap on LIBERO-Plus highlights robustness
differences that are less visible under the original evaluation
conditions.
Appendix~\ref{app:libero_plus} reports category and difficulty
breakdowns for the shared evaluation instances.

\subsection{Action-Centric Interaction}
\label{sec:interaction_analysis}
\label{sec:adaptation_analysis}

\textbf{Where Should the Streams Interact?}
Table~\ref{tab:interaction} compares cross-stream attention topologies within our architecture.
All variants allow action queries to read both semantic and dynamics information, so the comparison concerns how context is constructed. \methodname{} achieves the highest success rate.
Adding semantic-to-dynamics attention allows dynamics queries to read semantic tokens.
Further allowing dynamics queries to read action tokens enables bidirectional attention between the dynamics and action streams.
Allowing all three streams to attend to one another yields the lowest success rate.
The observed ordering is consistent with a benefit from avoiding additional cross-stream dependencies during context construction.
In particular, independently constructed semantics and dynamics can instead be weighted and combined by action queries as the action representation evolves.
Despite this forward independence, both context backbones remain
trainable through action supervision.
Appendix~\ref{app:interaction_controls} specifies the attention
masks and inference dependencies; Appendix~\ref{app:joint_adaptation}
reports freezing and prediction-only Cosmos controls.

\begin{wraptable}{r}{0.50\linewidth}
    \vspace{-0.4cm}
    \centering
    \caption{Attention-topology ablation on RoboCasa.
    Entries list the streams read by each query stream.
    The last row is \methodname{}.}
    \label{tab:interaction}
    \small
    \setlength{\tabcolsep}{3pt}
    \renewcommand{\arraystretch}{1.10}
    \begin{tabular*}{\linewidth}{@{\extracolsep{\fill}}cccc@{}}
        \toprule
        \Vtag{} reads & \Atag{} reads & \Wtag{} reads & SR (\%) \\
        \midrule
        \Vtag{}
            & \Vtag\,\Atag\,\Wtag
            & \Vtag\,\Wtag
            & 64.6 \\
        \Vtag{}
            & \Vtag\,\Atag\,\Wtag
            & \Vtag\,\Atag\,\Wtag
            & 64.0 \\
        \Vtag\,\Atag\,\Wtag
            & \Vtag\,\Atag\,\Wtag
            & \Vtag\,\Atag\,\Wtag
            & 59.1 \\
        \midrule
        \Vtag{}
            & \Vtag\,\Atag\,\Wtag
            & \Wtag{}
            & \textbf{71.2} \\
        \bottomrule
    \end{tabular*}
    \vspace{-6pt}
\end{wraptable}
\textbf{Which Context Layers Should Action Queries Read?}
We next examine the layerwise interleaving used by \methodname{}.
Each action layer reads semantic and dynamics keys and values (K/V)
from the corresponding VLM and WM layers.
Figure~\ref{fig:context_ablations}(a) compares this design with three
controls that reuse the final-layer K/V from the VLM, the WM, or
both at every action layer.
When final-layer reuse is applied to only one stream, the other
retains its original layerwise mapping.
All variants reuse the existing K/V projections and differ only
in which context layers action queries read. Figure~\ref{fig:context_ablations}(a) shows that \methodname{}
achieves 71.2\% success with layerwise access to both streams,
compared with 65.8\%, 63.5\%, and 64.6\% when final-layer K/V
reuse is applied to the VLM only, the WM only, or both streams, respectively.
A plausible explanation is that final-layer representations may not expose all the fine-grained motion cues useful for control.
Repeated final-layer attention reweights the same context, whereas layerwise interleaving provides direct access to intermediate semantic and dynamics features.
These features may help combine local control details with task-level guidance during action refinement.

\subsection{Generated Context and Dynamics Supervision}
\label{sec:context_analysis}
\label{sec:ablations}

\begin{wrapfigure}{r}{0.65\textwidth}
    \centering
    \vspace{-0.45cm}
    \includegraphics[
        height=0.351\linewidth
    ]{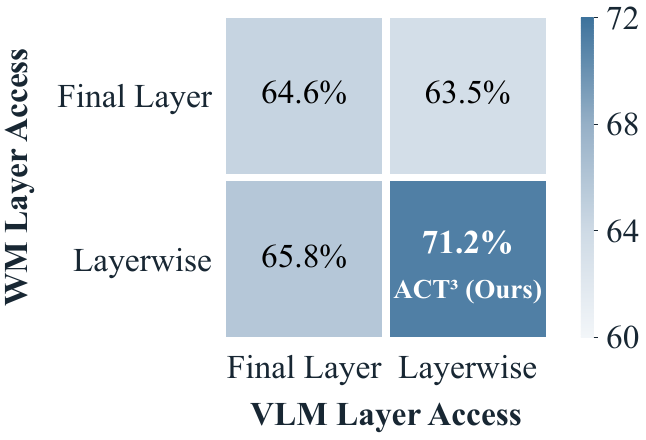}%
    \hspace{0.01\linewidth}%
    \includegraphics[
        height=0.351\linewidth
    ]{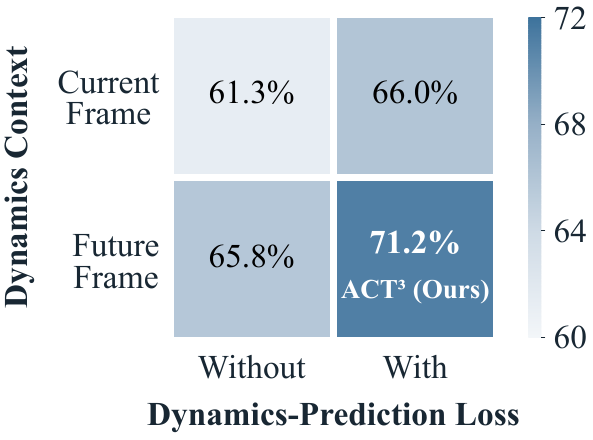}
    \caption{Effects of context-layer access and dynamics prediction
        on RoboCasa.
        \textbf{(a)~Left:} Context-layer access in the VLM and WM streams.
        ``Layerwise'' uses the original layer correspondence;
        ``Final Layer'' reuses the last layer's K/V at every action layer.
        \textbf{(b)~Right:} Separating future-context generation from
    dynamics-prediction supervision.
    Rows specify current-frame or generated-future dynamics context;
    columns indicate whether the prediction loss is applied.
    Each cell reports success rate (\%); the bottom-right cell
    in each panel corresponds to \methodname{}.}
    \label{fig:context_ablations}
    \vspace{-0.3cm}
\end{wrapfigure}
The dynamics stream contributes both context for action generation
and an auxiliary prediction objective.
Figure~\ref{fig:context_ablations}(b) separates these roles by
varying current-frame versus generated-future context, with and
without the dynamics-prediction loss.
All variants use the same architecture and update Cosmos through
action supervision.
In the current-frame setting, prediction supervision uses a
separate branch with shared dynamics parameters, so action queries
still receive only current-frame dynamics context. 
Generated-future context increases success from 61.3\% to 65.8\%
without prediction supervision and from 66.0\% to 71.2\% with it.
This gain may arise because action queries can read anticipated
local state changes directly, reducing the need to infer them
from current-frame features alone.
Prediction supervision also benefits current-frame control,
despite no generated future reaching the action stream.
This suggests that learning demonstrated transitions helps shape
control-relevant representations in the shared dynamics backbone,
beyond its role in training future generation.
Appendix~\ref{app:context_controls} details the computation paths
and factorial contrasts; Appendix~\ref{app:temporal_targets}
reports supplementary target-timing experiments.

\subsection{Analysis of Action-Centric Attention}
\label{sec:event_readout}

\begin{wrapfigure}{r}{0.61\linewidth}
    \centering
    \vspace{-12pt}
    \includegraphics[width=\linewidth]
    {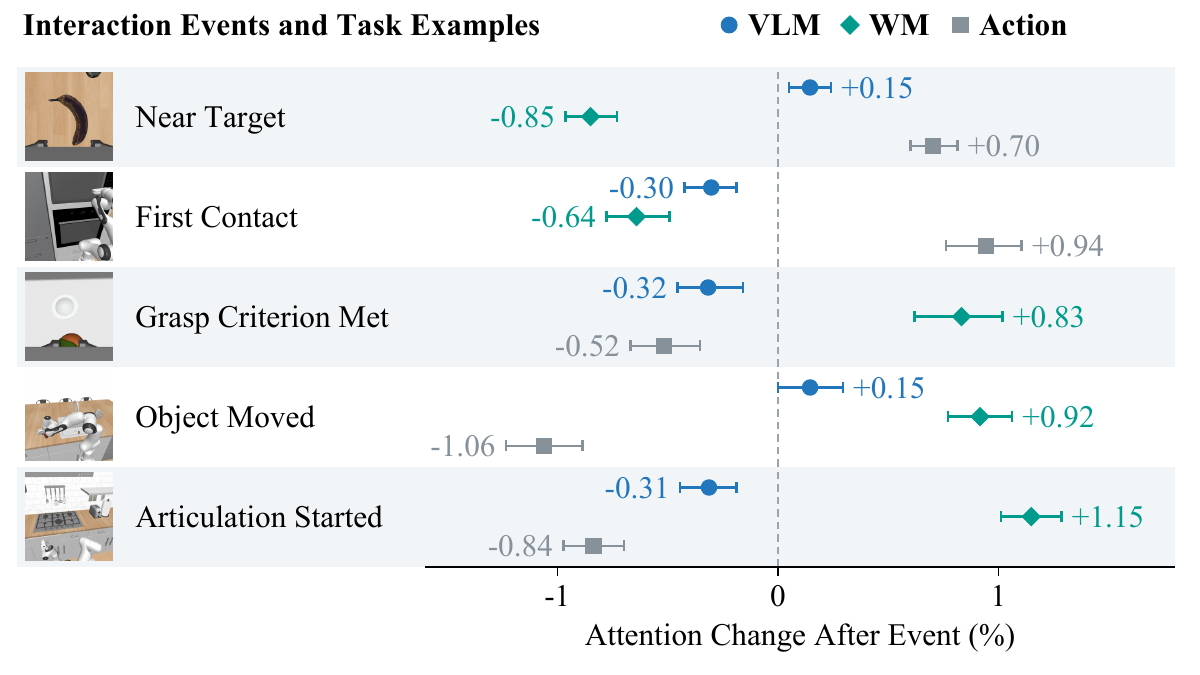}
    \vspace{-0.8cm}
    \caption{Event-dependent allocation of action-query attention.
    Points and adjacent numbers are task-equal mean post-minus-pre
    changes. Error bars show 95\% bootstrap intervals.}
    \label{fig:event_readout}
    \vspace{-8pt}
\end{wrapfigure}
We examine how action queries allocate attention across the independently
computed VLM context, WM context, and tokens of the action chunk during
interaction.
We collect 300 rollouts of \methodname{} across six RoboCasa tasks
and compare attention before and after five simulator-defined events. For each event, we first compute
a within-rollout change and then average across rollouts within each task,
giving contributing tasks equal weight.
Figure~\ref{fig:event_readout} reports these changes in percentage points;
positive values indicate increased attention after the event.
Appendix~\ref{app:attention_readout} provides the measurement protocol.
Near-target and first-contact events are associated with increased
attention to action tokens and decreased attention to WM context.
One possible interpretation is that approach and initial
contact place greater emphasis on local adjustments within the ongoing
action chunk. Around the grasp criterion, object motion, and articulation
onset, WM attention instead increases while action-token attention decreases.
These events involve sustained bilateral contact or changes in object
state, making the increased WM readout consistent with a greater need to
account for interaction dynamics.
VLM attention changes more modestly, possibly because the
task
goals and object roles remain largely unchanged across these short event
windows, sustaining a need for visual-language grounding. 

\subsection{Real-World Experiments}
\label{sec:real_world}

\textbf{Task Setup.}
We evaluate three real-world  robotic manipulation tasks on the AgileX CobotMagic bimanual
platform with one external and two wrist-mounted RGB cameras.
The tasks are \emph{Bowl Stacking}, \emph{Block Collection}, and
\emph{Flower Arrangement}, covering stable placement,
repeated object transfer, and ordered multi-stage execution, respectively.
Flower Arrangement requires pouring water into a vase before
inserting the flowers.
We train a separate policy for each task and method using 100, 100,
and 80 demonstrations, respectively.
Appendix~\ref{app:real_world} provides training details, evaluation
protocols, and success criteria.

\begin{wraptable}{r}{0.5\textwidth}
    \centering
    \vspace{-0.42cm}
    \caption{Average success rates (\%) in real-world manipulation tasks.
    Entries are successful trials / total trials (success rate).
    Success criteria are defined in
    Appendix~\ref{app:real_world_criteria}.}
    \label{tab:real_world_success}
    \small
    \setlength{\tabcolsep}{4pt}
    \begin{adjustbox}{max width=\linewidth}
    \begin{tabular}{@{}lcc@{}}
        \toprule
        Task & $\pi_{0.5}$ & \textbf{\boldmath\methodname}~(Ours) \\
        \midrule
        Bowl Stacking
            & 26/30 (86.7\%) & 27/30 (90.0\%) \\
        Block Collection
            & 22/30 (73.3\%) & 25/30 (83.3\%) \\
        Flower Arrangement
            & 14/30 (46.7\%) & 19/30 (63.3\%) \\
        \bottomrule
    \end{tabular}
    \end{adjustbox}
\end{wraptable}

\textbf{Task Completion.}
We evaluate each method for 30 trials per task, with initial object
configurations independently randomized across trials and methods
for all three tasks.
Table~\ref{tab:real_world_success} reports full-task success counts
and rates.
On the Bowl Stacking task, \methodname{} achieves a 90.0\% success rate, compared
with 86.7\% for $\pi_{0.5}$.
On the Block Collection task, \methodname{} achieves 83.3\%, compared with
73.3\% for $\pi_{0.5}$.
The largest observed gain occurs on the  Flower Arrangement task,
where \methodname{} achieves a 63.3\% success rate compared with 46.7\%
for $\pi_{0.5}$.

\begin{wrapfigure}{r}{0.70\linewidth}
    \centering
    \vspace{-12pt}
    \includegraphics[width=\linewidth]
    {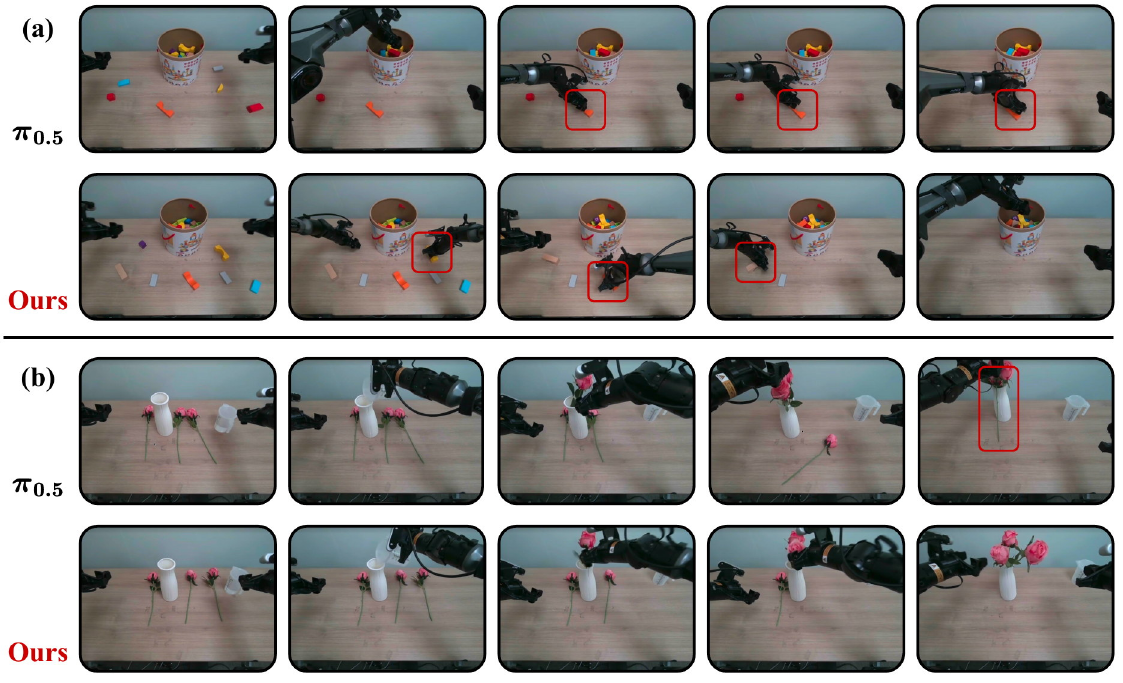}
    \caption{Qualitative comparison of real-world executions:
(a) Block Collection and (b) Flower Arrangement.
Each panel shows $\pi_{0.5}$ (top) and \methodname{} (bottom),
with keyframes ordered from left to right.
Red boxes highlight grasping and insertion details.}
    \label{fig:real_world_execution}
    \vspace{-8pt}
\end{wrapfigure}

\textbf{Execution Analysis.}
Figure~\ref{fig:real_world_execution} compares selected execution
sequences on the hard Block Collection and Flower Arrangement tasks.
In collection, $\pi_{0.5}$ repeatedly attempts to grasp an arch-shaped
block without attaining a suitable gripper orientation.
In contrast, our \methodname{} aligns its gripper with the target blocks
and successfully grasps several arch-shaped blocks in succession.
In the pouring-and-flower sequence, $\pi_{0.5}$ brings the final flower
toward the vase, but the stem passes alongside the rim instead of
entering the opening.
\methodname{} aligns the final stem with the opening and successfully
inserts it.
These cases illustrate how \methodname{} sustains task progress
by achieving the precise alignment required for grasping and insertion.

\section{Conclusion}

In this paper, we introduced \methodname{}, an Action-Centric
Tri-Stream Transformer that combines semantic understanding and
predictive dynamics for robot control.
The action expert serves as the sole integration point through
layerwise attention, while the semantic and dynamics streams
maintain independent forward computation.
Joint optimization of action generation and dynamics prediction
allows both context representations and their integration to
adapt to the control objective.
Experiments on simulated and real-world manipulation tasks demonstrate the effectiveness of this design.
The experimental findings suggest that global task semantics and
local predictive dynamics can be effectively combined through
action generation without directly coupling the two context streams.
Our current evaluation is limited to robotic manipulation;
future work will explore extending this action-centric design
to other embodied tasks, such as navigation.

\subsection*{AI Use Statement}

Generative AI tools were used to assist with polishing the manuscript and with code implementation. All AI-assisted text and code were reviewed and verified by the authors. The authors take full responsibility for the final content of this work.

\subsection*{Reproducibility statement}
We provide detailed formulations and experimental specifications to facilitate reproducibility. Our implementation builds on the OpenPI codebase. Section~\ref{sec:method} and Appendix~\ref{app:method_details} describe the architecture, training objectives, and inference procedure of \methodname{}. Appendix~\ref{app:exp_protocol} documents the datasets, model initialization, training configurations, and evaluation protocols, while Appendix~\ref{app:real_world} provides the corresponding real-world experimental details.

\bibliography{iclr2027_conference}
\bibliographystyle{iclr2027_conference}

\newpage
\appendix
\part*{Appendix}

\vspace*{20pt}
\section*{Table of Contents}
\hypersetup{
  linkcolor=darkblue   
}
\startcontents[sections]
\printcontents[sections]{l}{1}{\setcounter{tocdepth}{2}}
\hypersetup{
  linkcolor=red   
}

\newpage

\section{Additional Discussion of Multi-Stream Architectures}
\label{sec:additional_related_work}

We provide a more detailed discussion of the multi-stream architectures summarized in Fig.~\ref{fig:architecture_comparison}, focusing on how their designs differ from \methodname{}.

\textbf{Semantic--Dynamics Coupling} methods construct dynamics representations on top of semantic representations before they are consumed by the action pathway.
F1~\citep{lv2025f1} organizes understanding (\textit{i.e.}, semantic), generation (\textit{i.e.}, dynamics), and action experts through progressive attention: the generation expert predicts visual foresight conditioned on the understanding stream, while the action expert subsequently accesses both semantic and dynamics representations.
InternVLA-A1~\citep{cai2026internvla} follows a similar understanding--generation--action hierarchy, using a cumulative attention mask such that the generation expert predicts future visual latents from the semantic prefix and the action expert attends to both streams.
BagelVLA~\citep{hu2026bagelvla} further factorizes decision making into textual subtask prediction, visual keyframe generation, and action generation, with the predicted semantic plan guiding visual foresight before both are used for control.
These methods therefore share an explicit semantic-to-dynamics dependency: dynamics representations are constructed after reading semantic information.
In contrast, \methodname{} does not impose an S$\rightarrow$D hierarchy.
Its pretrained VLM and WM construct semantic and dynamics contexts independently, and their representations meet only through the action stream.
Thus, the action expert becomes the first point at which semantic and dynamics information is integrated.

\textbf{Dynamics--Action Coupling} methods directly connect predictive dynamics with the evolving action representation.
Motus~\citep{bi2026motus} combines semantic, video-generation, and action experts through joint attention and jointly models future video and actions, making the dynamics representation part of the same generative process as control.
DUST~\citep{won2026dust} similarly models future visual states and actions with two diffusion streams whose representations interact during denoising, while semantic features from a frozen VLM provide upstream conditioning.
STARRY~\citep{tian2026starry} uses a pretrained spatial-temporal world model together with semantic and action experts, and jointly denoises future spatial-temporal latents and actions through multimodal joint attention; its geometry-aware mechanism further modulates action-to-video interaction.
Although these designs differ substantially in implementation, they couple dynamics and action representations during joint generation.
\methodname{} instead adopts asymmetric information flow: only action queries read both semantic and dynamics contexts.
The two context hierarchies therefore remain action-independent during forward computation and can be cached throughout action denoising.
Importantly, this independence does not imply freezing: action gradients still update both pretrained context backbones, while the WM additionally receives dynamics-prediction supervision.

\textbf{Feature Conditioning.}
Feature-conditioning methods keep semantic and dynamics pathways separate and provide their outputs as upstream context for the action policy.
TriVLA~\citep{liu2025trivla} follows this design with a pretrained VLM for semantic grounding and a video-diffusion model for dynamics representations.
Multi-level video features are aggregated into dynamics tokens, which, together with VLM features, condition a downstream diffusion-transformer policy through cross-attention.
This design avoids direct S--D coupling and is therefore closest to \methodname{} in terms of independent context construction.
The key distinction lies in how these contexts are exposed to actions and adapted for control.
TriVLA first extracts and aggregates semantic and dynamics features and subsequently uses them as conditioning inputs, with both context modules frozen during downstream policy learning.
In \methodname{}, by contrast, each action layer directly accesses the corresponding VLM and WM keys and values, allowing the evolving action representation to draw on different depths of both pretrained representation hierarchies.
Moreover, both context backbones remain trainable through action supervision.

\section{Experimental Protocol and Reproducibility}
\label{app:exp_protocol}

\subsection{Datasets and Preprocessing}
\label{app:data}

\textbf{LIBERO.}
Following \citet{kim2024openvla}, we use LIBERO demonstration data
with unsuccessful episodes and no-op transitions removed.
We convert the training splits of LIBERO-Spatial, Object, Goal,
and Long into a single LeRobot dataset and train one policy jointly
on all 40 tasks.
The conversion preserves the episodes and frames in the preprocessed
training splits without applying an additional success filter.
The resulting dataset contains 1,693 episodes and 273,465 stored
frames, with 432 episodes in Spatial, 454 in Object, 428 in Goal,
and 379 in Long.
The nominal budget of 50 demonstrations per task refers to the
original dataset before filtering.

\textbf{RoboCasa.}
Following \citet{kim2026cosmospolicy}, we use the successful subset
of the human demonstration data for the 24 RoboCasa kitchen
manipulation tasks in the nominal 50-demonstration-per-task setting.
We convert this subset into a single LeRobot dataset containing
1,199 episodes and 316,995 stored frames.
The dataset includes 237 distinct instruction strings associated
with the 24 environment tasks.
Both action learning and dynamics-prediction supervision use
exclusively this set of successful demonstration trajectories.

\textbf{Camera observations.}
The LIBERO semantic stream receives an external view and a wrist view.
On RoboCasa it receives the left and right external views and the wrist
view. The dynamics stream uses only the designated base view: the
external camera on LIBERO and the left external camera on RoboCasa.
Images are resized with aspect-ratio-preserving padding to
$224\times224$ and mapped to $[-1,1]$.
During training, non-wrist images in the semantic branch undergo a random
crop with approximately 95\% of the original side length, resizing, and
rotation within $\pm5^{\circ}$; all views receive brightness, contrast,
and saturation factors in $[0.7,1.3]$, $[0.6,1.4]$, and $[0.5,1.5]$,
respectively. Augmentation parameters are sampled per camera and shared
within a batch. The current-and-future clip supplied to the auxiliary dynamics-supervision branch is resized
but receives no random augmentation. Consequently, the two branches do
not share an augmented current image during training.

\textbf{Actions and proprioception.}
Both datasets provide seven task-action channels for end-effector
translation, rotation, and gripper control. We use the stored action
convention without applying an additional delta-action transform.
The normalization maps each channel's training quantiles $q_{0.01}$ and
$q_{0.99}$ to approximately $-1$ and $1$, without clipping outliers.
Action targets are zero-padded to $B=32$ channels; the flow-matching
loss averages over all 32 channels and all action steps, without a
padding-channel loss mask. Execution retains only the seven task channels.
The data interface carries an eight-dimensional state on LIBERO and a
nine-dimensional state on RoboCasa. Proprioceptive input is disabled in
the main experiments: these states are not supplied to either policy.

\textbf{Temporal construction and boundaries.}
For an example starting at stored frame $t$, the action target is
$[a_t,\ldots,a_{t+9}]$ and the supervision clip contains images at
$[t,t+1,t+2,t+3,t+4]$. These are consecutive dataset indices.
The conversions assign metadata rates of 10\,Hz for LIBERO and 20\,Hz
for RoboCasa; the corresponding four-step spans on the converted clock
are 0.4\,s and 0.2\,s. They do not establish physical elapsed time after
frame filtering or the deployed control frequency.
LeRobot clamps out-of-range indices to the episode boundary. The policy
adapter supplies an all-valid future-frame mask rather than forwarding
the dataset's padding flags; repeated terminal frames therefore remain
in the prediction loss. This boundary handling is shared by the selected
context variants. Alternative offset sets are specified in
Appendix~\ref{app:temporal_targets}.

\subsection{Model Initialization and Gradient Paths}
\label{app:model}

\textbf{Checkpoints and update scope.}
We initialize the semantic and action streams from the pretrained
$\pi_{0.5}$ base model~\citep{intelligence2025pi} and the dynamics
stream from the post-trained Cosmos-Predict2.5-2B video generation
checkpoint~\citep{cosmos_predict25}.
We jointly optimize the PaliGemma backbone, including its vision
encoder, the action expert, the Cosmos transformer, and their
trainable projections.
The video VAE and text encoder remain frozen.
To match the $\pi_{0.5}$ attention interface, we reshape the
Cosmos Q/K/V tensors by merging adjacent heads before computing
attention in the interleaved context-extraction blocks.

\textbf{Video latents and action context.}
The frozen video VAE compresses a five-image $224\times224$ clip into
two $16$-channel $28\times28$ latent-time slices, corresponding to
the observed frame and the future segment.
Latent normalization follows the pretrained Cosmos model.
For current-frame conditioning, we encode the current image followed
by four zero-filled frames and keep the observed latent slice fixed.
Starting from fresh Gaussian noise in the future region, we generate
the future latent using two sampling steps conditioned on the current
frame and instruction, without classifier-free guidance.
The final future slice summarizes the temporally encoded future
segment and is embedded into 196 tokens using $(1,2,2)$ patches.
We process this slice as a clean latent input to the dynamics stream,
whose layerwise representations provide context for action generation.

\textbf{Supervision and gradient flow.}
In the default model, the supervised video branch corrupts future latents using
$\sigma\sim\mathcal{U}(0.001,0.999)$ and reconstructs clean latents from
the predicted velocity. The observed slice is fixed, with conditioning
timestep 0.1. The auxiliary loss is a float32 mean squared error over
future latent entries only. Its default weight is
$\lambda_{D}=1$.
The action branch constructs its future context using only the current
image and instruction, independently of the supervised branch. Its two-step sampler runs with autograd
enabled during training, with no detach between sampling and context
extraction. Thus, action-loss gradients reach the shared Cosmos parameters
through both sampling and subsequent feature extraction, as well as the
semantic context stream. In this default formulation, the demonstrated future is used only by the
supervised video branch. Appendix~\ref{app:privileged_distillation} describes
the training-only privileged teacher exception. Inference runs without gradients and does not access
future observations.

\subsection{Optimization and Checkpoint Selection}
\label{app:optimization}

Table~\ref{tab:implementation} summarizes the training recipe.
We optimize all trainable parameters with AdamW using a shared
learning rate of $5\times10^{-5}$ after linear warm-up, a global
batch size of 256, and no exponential moving average of model weights.
For action flow matching, we sample
$\tau=0.001+0.999u$, where $u\sim\mathrm{Beta}(1.5,1)$,
and draw action noise from a standard Gaussian distribution.
We train our method on eight NVIDIA H100 GPUs using Accelerate
with DeepSpeed ZeRO-2 and BF16 mixed precision; 10,000 optimizer
updates take approximately 23 hours on this hardware.
\begin{table}[t]
\centering
\caption{Selected main training configurations. Settings retained from the main-study recipe.}
\label{tab:implementation}
\small
\begin{tabular}{lll}
\toprule
Setting & LIBERO & RoboCasa \\
\midrule
Final optimizer updates & 30,000 & 20,000 \\
Global batch size & 256 & 256 \\
Warm-up updates & 10,000 & 10,000 \\
Learning rate after warm-up & $5\times10^{-5}$ & $5\times10^{-5}$ \\
AdamW $(\beta_1,\beta_2)$ & $(0.9,0.95)$ & $(0.9,0.95)$ \\
AdamW $\epsilon$ / weight decay & $10^{-8}$ / $10^{-10}$ & $10^{-8}$ / $10^{-10}$ \\
Gradient norm limit & 1.0 & 1.0 \\
Data-loader workers & 8 & 8 \\
EMA in PyTorch trainer & None & None \\
Action horizon $H$ & 10 & 10 \\
Proprioceptive input & Disabled & Disabled \\
Future images / video sampling steps & 4 / 2 & 4 / 2 \\
Action-flow integration steps & 10 & 10 \\
Dynamics-loss weight $\lambda_{D}$ & 1.0 & 1.0 \\
\bottomrule
\end{tabular}
\end{table}

The main RoboCasa study uses checkpoints after 20,000 updates,
while all LIBERO runs are trained for 30,000 updates.
The same LIBERO-trained checkpoints are evaluated on LIBERO-Plus
without additional fine-tuning.
Main-study checkpoints are saved every 10,000 updates, and we
evaluate the checkpoints at these prespecified endpoints.
The reported frozen-VLM and shifted-future ablations on RoboCasa
also use 20,000-update checkpoints.
Both policies start from the same pretrained VLA and use a matched
action-training protocol.

\subsection{Evaluation and Statistical Reporting}
\label{app:evaluation}

\textbf{Evaluation protocol.}
All in-house comparisons use matched evaluation conditions within
each benchmark. We report mean success rates (\%) over evaluation
rollouts, with each group score computed over the rollouts in that group.
The LIBERO and RoboCasa main experiments and ablations use three
evaluation seeds. For LIBERO, each seed includes 50 trials for each
of 40 tasks (6,000 trials in total). For RoboCasa, each seed includes
ten trials in each of five scenes for each of 24 tasks
(3,600 trials in total). On LIBERO-Plus, both policies are evaluated
on the same 10,030 perturbed instances.

\textbf{LIBERO implementation.}
The evaluator uses the benchmark's task-specific initial-state
list in index order.
The maximum action steps are 220, 280, 300, and 520 for Spatial,
Object, Goal, and Long, respectively.
Camera images are rotated by $180^{\circ}$ and resized/padded
to the policy resolution.

\textbf{RoboCasa implementation.}
We evaluate on object split B using five layout--style pairs,
$(1,1)$, $(2,2)$, $(4,4)$, $(6,9)$, and $(7,10)$, following the
Cosmos Policy~\citep{kim2026cosmospolicy} evaluation setup.
The trial allocation ensures equal coverage of these five scenes.
The evaluator uses the PandaMobile environment with three
camera views, $224\times224$ rendering, no camera randomization,
and vertical image flipping.
The seven predicted manipulator-action channels are extended
with the fixed base-control vector $[0,0,0,0,-1]$ before being
passed to the environment.
Table~\ref{tab:robocasa_tasks} lists the maximum episode length
for each task.

\begin{table}[t]
\centering
\caption{The 24 RoboCasa task identifiers used by the evaluator and
maximum action steps, excluding settling steps.}
\label{tab:robocasa_tasks}
\small
\begin{tabular}{lr@{\hspace{1.5em}}lr}
\toprule
Task & Steps & Task & Steps \\
\midrule
PnPCounterToCab & 500 & PnPCabToCounter & 500 \\
PnPCounterToSink & 700 & PnPSinkToCounter & 500 \\
PnPCounterToMicrowave & 600 & PnPMicrowaveToCounter & 500 \\
PnPCounterToStove & 500 & PnPStoveToCounter & 500 \\
OpenSingleDoor & 500 & CloseSingleDoor & 500 \\
OpenDoubleDoor & 1,000 & CloseDoubleDoor & 700 \\
OpenDrawer & 500 & CloseDrawer & 500 \\
TurnOnStove & 500 & TurnOffStove & 500 \\
TurnOnSinkFaucet & 500 & TurnOffSinkFaucet & 500 \\
TurnSinkSpout & 500 & CoffeeSetupMug & 600 \\
CoffeeServeMug & 600 & CoffeePressButton & 300 \\
TurnOnMicrowave & 500 & TurnOffMicrowave & 500 \\
\bottomrule
\end{tabular}
\end{table}

\textbf{Inference Latency.}
We measure model-query latency on a single NVIDIA H100 80GB GPU.
Each policy retains its native attention and precision settings;
action and sampling configurations follow
Table~\ref{tab:implementation}.
Mean latency on LIBERO/RoboCasa is 423/418\,ms for \methodname{},
compared with 161/160\,ms for $\pi_{0.5}$.
Timing includes future generation and context extraction but
excludes input/output transforms and data transfers.
For reference, Cosmos Policy~\citep{kim2026cosmospolicy}
reports 610\,ms per action chunk with five denoising steps on a
single H100 GPU, reduced to 160\,ms with one step.
These published timings provide context rather than a controlled
comparison.

\section{Architecture and Context-Control Details}

\subsection{Attention-Topology Ablations}
\label{app:interaction_controls}

\textbf{Block-level visibility.}
Rows index query streams and columns index key streams,
both ordered as $(S,A,D)$.
An entry of $0$ permits attention and $-\infty$ blocks it.
The four topologies in Table~\ref{tab:interaction} use
\begin{align}
\mathcal{M}_{\mathrm{ACT}}
&=\begin{pmatrix}
    0&-\infty&-\infty\\
    0&0&0\\
    -\infty&-\infty&0
\end{pmatrix},
&
\mathcal{M}_{\mathrm{SD}}
&=\begin{pmatrix}
    0&-\infty&-\infty\\
    0&0&0\\
    0&-\infty&0
\end{pmatrix},
\\
\mathcal{M}_{\mathrm{SD+AD}}
&=\begin{pmatrix}
    0&-\infty&-\infty\\
    0&0&0\\
    0&0&0
\end{pmatrix},
&
\mathcal{M}_{\mathrm{full}}
&=\begin{pmatrix}
    0&0&0\\
    0&0&0\\
    0&0&0
\end{pmatrix}.
\end{align}
Here, $\mathcal{M}_{\mathrm{ACT}}=\mathcal{M}$ in Eq.~\ref{eq:joint_attention} corresponds to \methodname{}.
Relative to this topology, $\mathcal{M}_{\mathrm{SD}}$ allows dynamics
queries to read semantic tokens, and $\mathcal{M}_{\mathrm{SD+AD}}$
further enables bidirectional attention between the dynamics
and action streams.
Under $\mathcal{M}_{\mathrm{full}}$, all three streams attend to one another.
These masks apply to the last 18 Cosmos blocks interleaved
with the semantic and action layers; the first ten blocks
retain independent computation.

\textbf{Training and inference.}
All variants jointly train the three transformer streams,
with the video VAE and Cosmos text encoder frozen.
Action-loss gradients reach both context backbones.
The two-step Cosmos sampler and separate dynamics-prediction
supervision branch remain unchanged.
The topology changes apply only to context extraction and
do not introduce joint denoising of future latents and actions.
At inference, one future latent is sampled per policy query
and held fixed throughout action-flow integration.
Under $\mathcal{M}_{\mathrm{ACT}}$ and $\mathcal{M}_{\mathrm{SD}}$, context features
are independent of action tokens, so their layerwise keys and
values are computed once per query and reused.
Under $\mathcal{M}_{\mathrm{SD+AD}}$, dynamics features depend on action
tokens; $\mathcal{M}_{\mathrm{full}}$ introduces this dependence into
semantic features as well.
Both feedback-enabled variants repeat joint feature computation
at each action-flow step while reusing the sampled future latent.

\subsection{Joint Adaptation of the Context Streams}
\label{app:joint_adaptation}

\textbf{Control definition.}
In the forward pass, the default model blocks cross-stream feedback
into the two context backbones, while allowing action-loss gradients
through the keys and values read by action queries during training.
The controls below retain this forward topology while restricting
parameter updates or their gradient sources.

\textbf{Frozen VLM.}
We freeze the entire PaliGemma backbone, including its vision
encoder, multimodal projector, and language model.
The action expert, its action and time projections, and the
Cosmos transformer remain trainable.
As shown in Table~\ref{tab:joint_adaptation}, this variant achieves
32.7\% success.
Freezing the VLM prevents its semantic representations from
co-adapting during training, leaving the action expert to combine
an unchanged semantic feature hierarchy with an evolving dynamics
stream.
Allowing the VLM to update instead lets action supervision reshape
these features as their integration with dynamics context is learned.
The low success suggests that effective integration benefits from
adapting the semantic context together with the action-side readout.
Independent forward computation therefore separates context
construction without requiring the semantic representations to
remain fixed.

\textbf{Frozen Cosmos.}
We freeze the entire Cosmos expert at its pretrained initialization,
while keeping the VLM and action expert trainable.
Generated future context and the action-centric attention topology
are retained, while dynamics-prediction supervision is disabled.
As shown in Table~\ref{tab:joint_adaptation}, this variant achieves
62.3\% success, compared with 71.2\% for the jointly adapted model.
Its improvement over the $\pi_{0.5}$ baseline
(Table~\ref{tab:robocasa_main}) suggests that the pretrained dynamics
stream provides useful context even when kept fixed.
However, freezing Cosmos prevents both future generation and the
intermediate representations read by actions from adapting to the
robot task. The other streams must therefore learn to use an
unchanged dynamics feature hierarchy.
Joint training instead allows predictive supervision to shape the
modeling of demonstrated transitions and action supervision to adjust
how this information supports control.
The observed ordering is consistent with benefiting from both
pretrained dynamics context and its task-specific adaptation.

\textbf{Prediction-only Cosmos.}
To isolate the contribution of action-driven Cosmos adaptation,
we keep Cosmos trainable but update it only through the
dynamics-prediction loss.
We stop action-loss gradients through the Cosmos keys and values
read by the action expert, while retaining generated-future context
and the same forward attention topology.
The VLM and action expert remain trainable, while the VAE and text
encoder remain frozen.
As shown in Table~\ref{tab:joint_adaptation}, this variant achieves
67.1\% success.
The improvement over frozen Cosmos suggests that adaptation to
demonstrated transitions benefits control even without action-driven
updates.
The further gain from joint adaptation supports the benefit of
action supervision beyond dynamics-prediction supervision alone:
action-loss gradients allow both future generation and the
intermediate representations read by action queries to adapt
directly to the control objective.

\begin{table}[t]
\centering
\caption{Success rates (\%) of context-backbone adaptation variants
on RoboCasa.
``Action'' and ``future'' denote the action loss and
dynamics-prediction loss, respectively.}
\label{tab:joint_adaptation}
\small
\setlength{\tabcolsep}{4pt}
\begin{tabular}{lllc}
\toprule
Variant & VLM updates & Cosmos Update Source & Average \\
\midrule
Frozen VLM & No & Action + Future & 32.7 \\
Frozen Cosmos & Yes & None & 62.3 \\
Prediction-only Cosmos & Yes & Future Only & 67.1 \\
\midrule
\textbf{\boldmath\methodname}~(Ours)
& Yes & Action + Future & 71.2 \\
\bottomrule
\end{tabular}
\end{table}

\subsection{Current and Generated Context}
\label{app:context_controls}

\textbf{Context construction.}
The four configurations in Figure~\ref{fig:context_ablations}(b)
use the same Cosmos backbone and action-context interface.
Current-frame variants select the observed latent slice, whereas
generated-future variants select the last temporal slice of the
sampled future latents.
In both cases, a single clean latent slice is embedded into 196
context tokens through the same interface.
The comparison therefore changes the source of dynamics context
without changing the context-token budget available to the action expert.

\textbf{Supervision and gradient paths.}
With current-frame context, the variant without dynamics-prediction
supervision skips both the auxiliary prediction branch and future
sampling.
Adding prediction supervision introduces a separate branch with
shared dynamics parameters, while action queries still receive
only current-frame dynamics context.
Neither current-frame variant performs future sampling at deployment.
With generated-future context, the variant without prediction
supervision retains differentiable future generation and action-loss
updates to Cosmos, but omits the auxiliary prediction branch.
The full \methodname{} model retains both differentiable future
generation and dynamics-prediction supervision.
Cosmos remains trainable through action supervision in all four
configurations; removing the dynamics-prediction loss does not
freeze the dynamics backbone.

\textbf{Factorial contrasts.}
Let $s_{c,\ell}$ denote success rate in percent, where
$c\in\{\mathrm{cur},\mathrm{gen}\}$ specifies current-frame or
generated-future dynamics context, and $\ell\in\{0,1\}$ indicates
whether dynamics-prediction supervision is used.
Both generated-future context and prediction supervision improve
success when the other factor is held fixed
(Figure~\ref{fig:context_ablations}(b)).
The interaction contrast measures how the gain from generated-future
context changes with prediction supervision:
\begin{equation}
\begin{aligned}
\Delta_{\mathrm{int}}
&= (s_{\mathrm{gen},1}-s_{\mathrm{cur},1})
 - (s_{\mathrm{gen},0}-s_{\mathrm{cur},0}) \\
&= (71.2-66.0)-(65.8-61.3)=0.7.
\end{aligned}
\label{eq:context_interaction}
\end{equation}
The point estimates suggest approximately additive benefits.
Generated-future context exposes anticipated local state changes
to action queries, while dynamics-prediction supervision may
improve control-relevant representations even with current-frame
context.

\subsection{Sampling, Supervision, and Action Integration}
\label{app:method_details}

\textbf{Future-context sampling.}
Let $E_{\mathrm{vae}}$ denote the frozen video VAE encoder,
including its pretrained latent normalization. We encode the base-camera
image followed by $N$ zero-filled frames as
\begin{equation}
    z_t^{\mathrm{cond}}
    =E_{\mathrm{vae}}([I_t^b,0,\ldots,0]).
    \label{eq:video_condition}
\end{equation}
A binary mask $\mathsf{M}_{\mathrm{obs}}$, with the same shape as
$z_t^{\mathrm{cond}}$, selects entries assigned to the observed frame
according to the VAE's temporal encoding; $1-\mathsf{M}_{\mathrm{obs}}$ selects the
unobserved future region.
For standard Gaussian noise $\epsilon_D$ of the same shape, the sampler
runs for $S_D$ steps:
\begin{equation}
    \begin{aligned}
        \widetilde z_t^{(0)}
        &=\mathsf{M}_{\mathrm{obs}}\odot z_t^{\mathrm{cond}}
          +(1-\mathsf{M}_{\mathrm{obs}})\odot\epsilon_D,\\
        \widetilde z_t^{(s+1)}
        &=\operatorname{Step}_{\phi}
          \left(\widetilde z_t^{(s)};
          z_t^{\mathrm{cond}},\ell,\mathsf{M}_{\mathrm{obs}},s\right),
          \quad s=0,\ldots,S_D-1.
    \end{aligned}
    \label{eq:video_rollout}
\end{equation}
Here $\odot$ denotes elementwise multiplication and
$\operatorname{Step}_{\phi}$ denotes one sampling update of the dynamics
model, including its noise schedule and restoration of the observed
region to $z_t^{\mathrm{cond}}$.
The instruction is supplied through the dynamics model's text encoder.
The default uses $N=4$ and $S_D=2$, without classifier-free guidance.
With this clip length, the VAE produces one observed and one future
latent-time slice; the latter represents the temporally encoded future
segment rather than a single future image.
After sampling, we treat the future slice of
$\widetilde z_t^{(S_D)}$ as a clean latent input and apply the dynamics
patch embedding to obtain dynamics tokens.
A subsequent transformer pass extracts the layerwise dynamics context $\mathcal{C}_{D,t}$
read by actions.
Sampling uses fresh noise independent of the demonstration targets.
The temporal offsets and episode-boundary handling follow
Appendix~\ref{app:data}; alternative target offsets are examined in
Appendix~\ref{app:temporal_targets}.

\textbf{Auxiliary dynamics supervision.}
The target clip is $Y_t=[I_t^b,I_{t+1}^b,\ldots,I_{t+N}^b]$, with
clean latent $z_0=E_{\mathrm{vae}}(Y_t)$.
We draw independent standard Gaussian noise $\epsilon_{\mathrm{sup}}$ of the same
shape and a video noise level
$\sigma\sim\mathcal{U}(\sigma_{\min},\sigma_{\max})$, where
$\mathcal{U}$ is the uniform distribution and
$0<\sigma_{\min}<\sigma_{\max}\leq1$.
Using the notation of Section~\ref{sec:joint_learning} at a fixed control step $t$, the intermediate corrupted target and conditioned input are
\begin{equation}
    \begin{aligned}
        \bar z_\sigma
        &=(1-\sigma)z_0+\sigma\epsilon_{\mathrm{sup}},\\
        z_\sigma
        &=\mathsf{M}_{\mathrm{obs}}\odot z_t^{\mathrm{cond}}
          +(1-\mathsf{M}_{\mathrm{obs}})\odot \bar z_\sigma.
    \end{aligned}
    \label{eq:video_noising}
\end{equation}
The noise-level field assigns a fixed conditioning level
$\sigma_{\mathrm{cond}}$ to observed entries and $\sigma$ to future
entries:
\begin{equation}
    \boldsymbol{\sigma}
    =\mathsf{M}_{\mathrm{obs}}\,\sigma_{\mathrm{cond}}
     +(1-\mathsf{M}_{\mathrm{obs}})\,\sigma.
    \label{eq:video_noise_field}
\end{equation}
The dynamics transformer $g_{\phi}$ predicts latent velocity from the
conditioned latent, noise-level field, instruction embedding, and mask.
Its clean-latent estimate is
\begin{equation}
    \hat z_0
    =\mathsf{M}_{\mathrm{obs}}\odot z_t^{\mathrm{cond}}
    +(1-\mathsf{M}_{\mathrm{obs}})\odot\left[
        \bar z_\sigma
        -\sigma g_{\phi}
        (z_\sigma,\boldsymbol{\sigma},\ell,\mathsf{M}_{\mathrm{obs}})
    \right].
    \label{eq:video_clean_estimate}
\end{equation}
As in sampling, the instruction argument denotes its text-encoder
embedding. We supervise only future entries:
\begin{equation}
    \mathcal{L}_{D}
    =\mathbb{E}_{\mathcal{D},\sigma,\epsilon_{\mathrm{sup}}}
    \left[
        \frac{
            \left\|(1-\mathsf{M}_{\mathrm{obs}})\odot
            (\hat z_0-z_0)\right\|_F^2
        }{\sum_j(1-(\mathsf{M}_{\mathrm{obs}})_j)}
    \right],
    \label{eq:video_loss}
\end{equation}
where $j$ indexes latent entries and the denominator counts future
entries.
Future sampling, context extraction, and auxiliary dynamics prediction
share the dynamics-stream parameters $\phi$ but use separate forward
passes.
Only sampling and context extraction provide dynamics inputs to the
action expert; demonstrated-future latents are used only by the
auxiliary branch in the default policy.
During training, action-loss gradients reach $\phi$ through both
sampling and context extraction, with no detach between them, while
$\mathcal{L}_{D}$ updates only the dynamics stream.
The semantic and action streams are updated by the action loss;
the video VAE and dynamics text encoder remain frozen.
Inference retains future sampling and context extraction but omits
the demonstrated-future supervision branch.

\textbf{Action-flow integration.}
For a fixed policy query, let $S_A$ be the number of Euler steps,
$\Delta\tau=1/S_A$ the step size, and $\tau_j=1-j\Delta\tau$ for
$j=0,\ldots,S_A$ the integration times.
Starting from a standard Gaussian $X_1\in\mathbb{R}^{H\times B}$,
we reuse the cached contexts and update
\begin{equation}
    X_{\tau_{j+1}}
    =X_{\tau_j}-\Delta\tau\,
      f_{\theta}\left(
          X_{\tau_j},\tau_j;
          \mathcal{C}_{S,t},\mathcal{C}_{D,t}
      \right),
    \quad j=0,\ldots,S_A-1.
    \label{eq:cached_euler}
\end{equation}
The final state $\hat A_t=X_0$ is the predicted action chunk.
The default uses $H=10$, $B=32$, and $S_A=10$.
Context caches remain fixed throughout these steps and are refreshed
for the next policy query.

\section{Supplementary Training Studies}
\label{app:supplementary_training}

\subsection{Temporal Sampling of Future Targets}
\label{app:temporal_targets}

Let $\mathcal{D}_f=\{d_1,d_2,d_3,d_4\}$ denote offsets in stored dataset
indices relative to observation $t$. The supervised clip is
\begin{equation}
Y_t(\mathcal{D}_f)
=[I_t^b,I_{t+d_1}^b,I_{t+d_2}^b,I_{t+d_3}^b,I_{t+d_4}^b].
\label{eq:offset_clip}
\end{equation}
We compare three temporal selections: consecutive near-future targets
$\{1,2,3,4\}$, consecutive later targets $\{7,8,9,10\}$, and
approximately uniform targets $\{2,5,7,10\}$ spread across a longer
future window. All settings contain four future images, while the
action target remains $[a_t,\ldots,a_{t+9}]$ and the execution prefix
remains five actions. This comparison varies temporal coverage while
keeping the number of predicted images fixed.

The selected images occupy the same four future slots of a five-image
clip. The VAE latent layout, two-step future sampler, and single-slice
action-context interface are retained. The actual dataset offsets are
not supplied as additional positional inputs to the dynamics model, so target
timing is specified through the training data. Offsets refer to stored
dataset indices; their relation to physical time and episode-boundary
clamping are described in Appendix~\ref{app:data}. Longer offsets can
produce more repeated terminal frames near episode ends.

\begin{table}[t]
\centering
\caption{RoboCasa success rates for different temporal selections
of future targets.}
\label{tab:future_offsets}
\small
\begin{tabular}{lcc}
\toprule
Target selection & Future offsets & SR (\%) \\
\midrule
Later consecutive & $\{7,8,9,10\}$ & 65.4 \\
Approximately uniform & $\{2,5,7,10\}$ & 62.5 \\
\midrule
\textbf{\boldmath\methodname}~(Ours)
& $\{1,2,3,4\}$ & 71.2 \\
\bottomrule
\end{tabular}
\end{table}

Table~\ref{tab:future_offsets} shows that consecutive near-future
targets achieve the highest success rate, while both later
consecutive and approximately uniform targets yield lower performance.
A plausible explanation is the relevance of nearby transitions to
receding-horizon control: the policy executes only a short prefix
of each action chunk before refreshing its observations and dynamics
context. Near-future supervision concentrates the fixed four-frame
budget on scene changes surrounding this next control segment.
Later consecutive targets shift that budget toward more distant
states, while approximately uniform targets spread it across a
longer interval with less local temporal detail.
The lower success of the uniform selection further suggests that
wider coverage does not compensate for losing densely sampled
local transitions.
Together, these results suggest that the temporal relevance and
local detail of dynamics context matter more than simply extending
its coverage in this setting.

\subsection{Exploratory Privileged Self-Distillation}
\label{app:privileged_distillation}

Our standard training objective uses demonstrated future frames to
supervise dynamics prediction. We investigate whether these same
observations can also provide additional guidance at the action level,
making fuller use of the future information available during training.
To this end, a privileged teacher branch predicts actions from
demonstrated-future latents encoded by the frozen video VAE, while
the student uses generated future context. The two branches share parameters and differ only in their future context.

We retain the original action and dynamics-prediction objectives and introduce
teacher action supervision and a distillation loss:
\begin{equation}
\begin{aligned}
\mathcal{L}_{\mathrm{KD}}
&=
\mathbb{E}\!\left[
\frac{
\|\widehat U^s_\tau-\operatorname{sg}(\widehat U^t_\tau)\|_F^2
}{HB}
\right], \\
\mathcal{L}_{\mathrm{priv}}
&=
\mathcal{L}_{A}^s
+0.25\,\mathcal{L}_{A}^t
+\lambda_{D}\mathcal{L}_{D}
+\lambda_{\mathrm{KD}}r(n)\mathcal{L}_{\mathrm{KD}}.
\end{aligned}
\label{eq:privileged_objective}
\end{equation}
Here, $\widehat U^s_\tau$ and $\widehat U^t_\tau$ are the student and
teacher action-flow predictions, and $\operatorname{sg}$ denotes
stop-gradient. The losses $\mathcal{L}_{A}^s$ and
$\mathcal{L}_{A}^t$ apply the same action-flow supervision
to the respective branches. The teacher action loss remains
differentiable and updates the shared parameters.
We test $\lambda_{\mathrm{KD}}\in\{0.1,0.5\}$.
For optimizer-update index $n$, the schedule $r(n)$ is zero before 5K updates, increases
linearly to one at 10K updates, and remains one thereafter.
Only the generated-future student path is used at deployment.

\begin{table}[t]
\centering
\caption{RoboCasa success rate (\%) with and without privileged
self-distillation.}
\label{tab:privileged_distillation}
\small
\begin{tabular}{lc}
\toprule
Training variant & SR (\%) \\
\midrule
Privileged KD, $\lambda_{\mathrm{KD}}=0.1$ & 65.0 \\
Privileged KD, $\lambda_{\mathrm{KD}}=0.5$ & 63.3 \\
\midrule
\textbf{\boldmath\methodname}~(Ours) & 71.2 \\
\bottomrule
\end{tabular}
\end{table}

Table~\ref{tab:privileged_distillation} shows that neither privileged
self-distillation variant improves on the \methodname{} reference,
with a lower success rate at $\lambda_{\mathrm{KD}}=0.5$ than at $0.1$.
One possible explanation is that the shared-parameter teacher provides
limited complementary guidance, since the original action objective
already adapts the dynamics stream through generated future context.
Moreover, the teacher observes the realized demonstrated future,
whereas the student relies on a future sampled from the current
observation and instruction. Matching their predictions may therefore
introduce competing supervision when these contexts support different
actions. 

\subsection{Details of Action-Centric Attention Analysis}
\label{app:attention_readout}

\textbf{Tasks and rollouts.}
We select six RoboCasa tasks covering object transfer and
articulated-object manipulation.
The object-transfer tasks are PnPCounterToSink,
PnPSinkToCounter, and PnPCounterToCab;
the articulated-object tasks are OpenSingleDoor,
OpenDrawer, and TurnOnStove.
Using the 20K RoboCasa checkpoint of \methodname{}, we collect ten
rollouts for each of five layout--style pairs per task.
The resulting set contains 300 rollouts and 21,307 policy queries,
including 210 successful and 90 unsuccessful rollouts.
Both outcomes are included in the analysis.

\textbf{Interaction events.}
Events are defined from simulator state independently of attention.
Near-target denotes the first end-effector distance of at most
0.10\,m from the object root or the resolved handle/knob centers.
First contact is detected using the gripper--target contact predicate.
The grasp criterion requires bilateral finger contact in two
consecutive control states.
Object motion denotes the first object displacement of at least
0.03\,m and is evaluated for the three object-transfer tasks.
Articulation onset denotes the first normalized joint-progress change
of at least 0.05 for a door or drawer, or a target-knob angle change
of at least 0.05\,rad for the stove.
\textbf{Attention measurement and aggregation.}
We measure attention at the final action-flow model evaluation
($\tau \simeq 0.1$), averaging over 18 layers, eight heads, and
action positions 0--4.
For each action query, attention probabilities are summed over the
visible keys belonging to VLM context, action tokens, and WM context,
yielding $m_S+m_A+m_D=1$.
VLM mass combines all three image streams and valid text-prefix tokens.
Policy queries are aligned within $\pm25$ control steps of each event
using five-step bins.
For each rollout with observations on both sides of an event,
we compute the change from its mean attention mass in the bins
centered at $-10$ and $-5$ to its mean in the bins centered at
$+5$ and $+10$.
We average these paired changes within each task and then weight
contributing tasks equally.
The 95\% confidence intervals are computed using 1,000 bootstrap
resamples of whole rollouts within each task.

\section{LIBERO-Plus Evaluation and Detailed Results}
\label{app:libero_plus}

\textbf{Perturbation categories.}
Table~\ref{tab:libero_plus_perturbations} reports results across
the seven perturbation categories.
\methodname{} improves over the baseline in every category, with
the largest gains for robot initial states and sensor noise.
These results indicate improved tolerance to both altered starting
configurations and degraded observations, extending the benefit
beyond changes in scene appearance.
Language-instruction perturbations also show a clear improvement,
demonstrating that the gain is not confined to visual changes.
Background textures, lighting conditions, and object layouts show
smaller improvements, with the baseline already performing well
in these categories.
Performance also improves under camera viewpoint changes, although
this remains the lowest-scoring category for both policies,
suggesting that changes in observation geometry remain more
demanding than texture or illumination changes in this evaluation.

\begin{table}[t]
    \centering
    \caption{LIBERO-Plus success rate (\%) by perturbation category.}
    \label{tab:libero_plus_perturbations}
    \small
    \begin{tabular}{lrrr}
        \toprule
        Perturbation category & Instances & $\pi_{0.5}$ & \textbf{\boldmath\methodname}~(Ours) \\
        \midrule
        Background textures & 1,076 & 87.6 & 91.3 \\
        Camera viewpoints & 1,599 & 47.1 & 58.5 \\
        Language instructions & 1,537 & 72.0 & 86.9 \\
        Lighting conditions & 1,142 & 93.7 & 94.8 \\
        Object layouts & 1,525 & 82.8 & 87.5 \\
        Robot initial states & 1,550 & 52.8 & 81.5 \\
        Sensor noise & 1,601 & 56.5 & 83.0 \\
        \midrule
        Overall & 10,030 & 68.4 & 82.4 \\
        \bottomrule
    \end{tabular}
\end{table}

\textbf{Difficulty levels.}
Table~\ref{tab:libero_plus_difficulty} groups instances by their
benchmark-provided difficulty annotations.
\methodname{} outperforms the baseline at every annotated level.
Although success rates decrease with difficulty for both policies,
the advantage of \methodname{} persists through the most difficult
group.
This pattern indicates that the robustness gains extend across
the evaluated difficulty range, rather than being confined to
easier instances.

\begin{table}[t]
    \centering
    \caption{LIBERO-Plus success rate (\%) by difficulty level.}
    \label{tab:libero_plus_difficulty}
    \small
    \begin{tabular}{lrrr}
        \toprule
        Difficulty level & Instances & $\pi_{0.5}$ & \textbf{\boldmath\methodname}~(Ours) \\
        \midrule
        1 & 1,644 & 83.2 & 94.6 \\
        2 & 2,202 & 77.8 & 92.7 \\
        3 & 2,094 & 72.7 & 87.2 \\
        4 & 1,886 & 63.6 & 78.0 \\
        5 & 2,083 & 45.4 & 59.9 \\
        Unannotated & 121 & 92.6 & 96.7 \\
        \midrule
        Overall & 10,030 & 68.4 & 82.4 \\
        \bottomrule
    \end{tabular}
\end{table}

\section{Real-World Experiment Details}
\label{app:real_world}

\subsection{Setup and Training}
\label{app:real_world_data}
\label{app:real_world_training}

The AgileX CobotMagic platform has six joints and one gripper per arm.
One top-down and two wrist-mounted RGB cameras record
$640\times480$ images at 30\,Hz.
We train a separate policy for each task and method using the
demonstration budgets in Section~\ref{sec:real_world}.
Training uses 40,000 optimizer updates, a global batch size of 64,
and an action horizon of $H=50$.
The learning rate increases linearly to $2.5\times10^{-5}$ over
1,000 warm-up updates, then follows a cosine decay to
$2.5\times10^{-6}$.
Other optimizer settings follow Table~\ref{tab:implementation}.

\subsection{Evaluation Protocol}
\label{app:real_world_protocol}

We evaluate $\pi_{0.5}$ and \methodname{} for 30 trials per task.
For all three tasks, initial object configurations are independently
randomized across trials and methods.
Each trial is scored using the task-specific success criteria
in Appendix~\ref{app:real_world_criteria}.
Table~\ref{tab:real_world_success} reports the number of successful
trials and the corresponding success rate, computed as the fraction
of successful trials out of 30 and expressed as a percentage.

\subsection{Tasks and Metrics}
\label{app:real_world_criteria}

\textbf{Bowl Stacking.}
All designated bowls must form the target stack after gripper release
and remain stable throughout the final observation interval.
The initial base bowl, when present, is excluded from the required
placements and the partial-completion count.

\textbf{Block Collection.}
All designated blocks must be fully inside the target bucket after
release and remain there throughout the final observation interval.
Partial completion counts the blocks retained in the bucket at
termination, not the number of transfer attempts.

\textbf{Flower Arrangement.}
The required order is to finish pouring before inserting flowers.
The cumulative milestones are (1) verified water transfer into the
vase and completed pouring, (2) at least one flower stem entering
the vase after pouring, and (3) all target flowers inserted, released,
and stable, with the vase upright.
Insertion before pouring is an ordering violation and receives no
credit for later milestones.
Full-task success requires reaching the third milestone.

\begin{figure}[htbp]
    \centering
    \begin{minipage}[t]{0.32\linewidth}
        \centering
        {\small\textbf{(a) Bowl Stacking}}\par
        \vspace{5pt}

        {\footnotesize
        \textcolor[HTML]{326C97}{\textbf{Initial state}}}\par
        \vspace{2pt}
        \includegraphics[width=\linewidth]
            {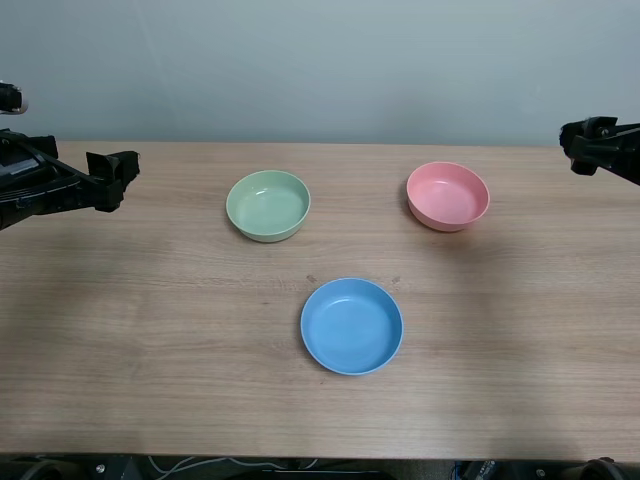}\par
        \vspace{5pt}

        {\footnotesize
        \textcolor[HTML]{277C66}{\textbf{Final state}}}\par
        \vspace{2pt}
        \includegraphics[width=\linewidth]
            {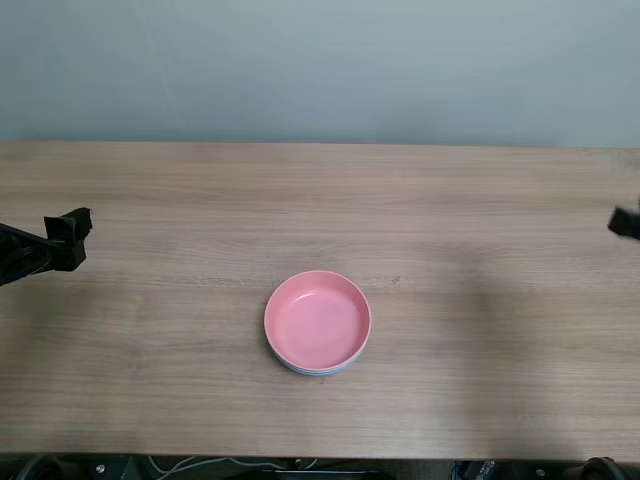}\par
        \vspace{5pt}

        {\footnotesize\raggedright
        \textbf{Goal.} Stack all designated bowls into a single stable
        stack that remains in place after gripper release.\par}
    \end{minipage}\hfill
    \begin{minipage}[t]{0.32\linewidth}
        \centering
        {\small\textbf{(b) Block Collection}}\par
        \vspace{5pt}

        {\footnotesize
        \textcolor[HTML]{326C97}{\textbf{Initial state}}}\par
        \vspace{2pt}
        \includegraphics[width=\linewidth]
            {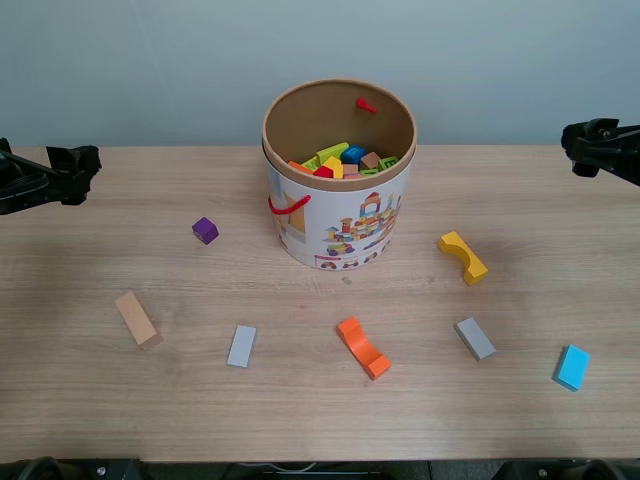}\par
        \vspace{5pt}

        {\footnotesize
        \textcolor[HTML]{277C66}{\textbf{Final state}}}\par
        \vspace{2pt}
        \includegraphics[width=\linewidth]
            {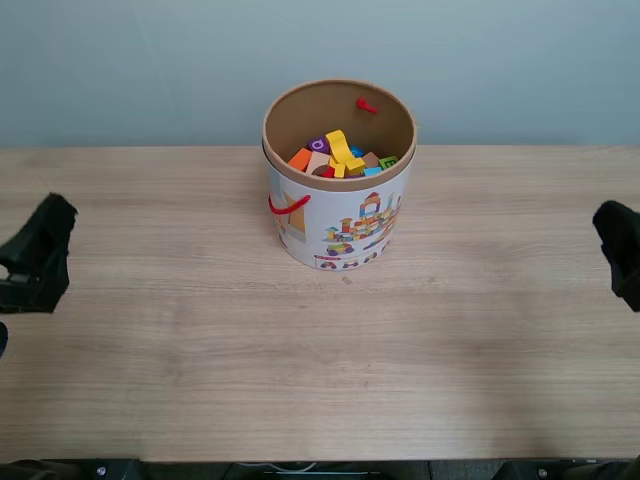}\par
        \vspace{5pt}

        {\footnotesize\raggedright
        \textbf{Goal.} Transfer all designated blocks from the tabletop
        into the bucket, leaving them fully inside after release.\par}
    \end{minipage}\hfill
    \begin{minipage}[t]{0.32\linewidth}
        \centering
        {\small\textbf{(c) Flower Arrangement}}\par
        \vspace{5pt}

        {\footnotesize
        \textcolor[HTML]{326C97}{\textbf{Initial state}}}\par
        \vspace{2pt}
        \includegraphics[width=\linewidth]
            {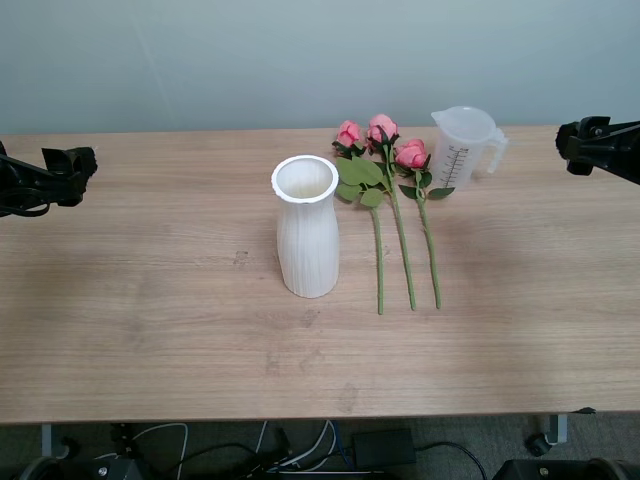}\par
        \vspace{5pt}

        {\footnotesize
        \textcolor[HTML]{277C66}{\textbf{Final state}}}\par
        \vspace{2pt}
        \includegraphics[width=\linewidth]
            {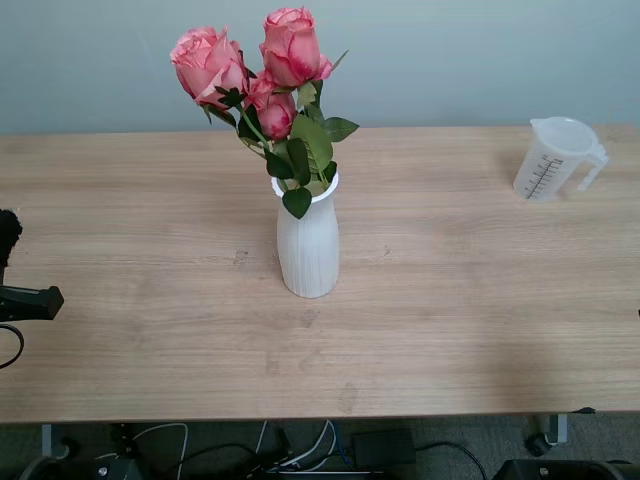}\par
        \vspace{5pt}

        {\footnotesize\raggedright
        \textbf{Goal.} Pour water into the vase, then insert and release
        all designated flowers, leaving the vase upright.\par}
    \end{minipage}

    \caption{Real-world task illustrations.
    Each column shows the initial state (top) and final state (bottom)
    from the same recorded episode.
    The final states show stacked bowls, collected blocks, and flowers
    retained in an upright vase, respectively.
    The intermediate water-transfer stage of Flower Arrangement is
    not visible in these endpoint images.}
    \label{fig:real_world_task_states}
\end{figure}

Figure~\ref{fig:real_world_task_states} illustrates the three real-world
tasks through paired initial and final observations.

\end{document}